\documentclass[11pt,a4paper]{article}
\usepackage[hyperref]{emnlp2020}
\usepackage{times}
\usepackage{latexsym}

\usepackage{microtype}

\usepackage{url}
\usepackage{graphicx}
\usepackage{wrapfig}
\usepackage{amsmath}
\usepackage{amssymb}
\usepackage{color,xcolor,colortbl}
\usepackage{enumitem}
\usepackage{algorithm}
\usepackage{algorithmic}
\usepackage[font={small}]{caption}
\usepackage{bm,bbm}
\usepackage{booktabs}
\usepackage{mathtools}
\usepackage{array}
\usepackage{multirow}
\usepackage{soul}
\usepackage{subcaption}
\usepackage{pbox}
\usepackage{pifont}
\usepackage{arydshln}

\newcommand\Tstrut{\rule{0pt}{2.6ex}}         

\newcommand\numberthis{\addtocounter{equation}{1}\tag{\theequation}}

\definecolor{darkblue}{rgb}{0.0, 0.0, 0.55}
\setul{0.5ex}{0.3ex} 
\setulcolor{blue} 
\definecolor{darkblue}{rgb}{0.0, 0.0, 0.55}
\definecolor{midnightblue}{HTML}{191970}
\definecolor{darkgreen}{HTML}{006400}
\definecolor{red}{HTML}{ff0000}
\definecolor{gold}{HTML}{ffd700}
\definecolor{mediumvioletred}{HTML}{c71585}
\definecolor{lime}{HTML}{00ff00}
\definecolor{aqua}{HTML}{00ffff}
\definecolor{fuchsia}{HTML}{ff00ff}
\definecolor{lightpink}{HTML}{ffb6c1}
\definecolor{dodgerblue}{HTML}{1e90ff}
\definecolor{deepskyblue}{HTML}{00bfff}
\definecolor{deeppink}{HTML}{ff1493}
\definecolor{orangered}{HTML}{ff4500}
\definecolor{mediumseagreen}{HTML}{3cb371}
\definecolor{saddlebrown}{HTML}{8b4513}
\DeclareRobustCommand{\legendsquare}[1]{\textcolor{#1}{\rule{3ex}{1.5ex}}}

\usepackage[T1]{fontenc}
\usepackage[scaled=.8]{beramono}

\aclfinalcopy 
\def\aclpaperid{3389} 

\title{Better Highlighting: Creating Sub-Sentence Summary Highlights}

\author{Sangwoo Cho,$^\dagger$ Kaiqiang Song,$^\dagger$ Chen Li,$^\S$, Dong Yu$^\S$, Hassan Foroosh$^\dagger$, Fei Liu$^\dagger$\\[0.8em]
$^\dagger$Computer Science Department, University of Central Florida\\
$^\S$Tencent AI Lab, Bellevue, WA, USA\\[0.6em]
\texttt{\{swcho,kqsong\}@knights.ucf.edu \quad \{ailabchenli,dyu\}@tencent.com}\\
\texttt{\{foroosh,feiliu\}@cs.ucf.edu}
}

\date{}

\begin{document}
\maketitle
\begin{abstract}

Amongst the best means to summarize is \emph{highlighting}.
In this paper, we aim to generate summary highlights to be overlaid on the original documents to make it easier for readers to sift through a large amount of text.
The method allows summaries to be understood in context to prevent a summarizer from distorting the original meaning, of which abstractive summarizers usually fall short.
In particular, we present a new method to produce \emph{self-contained} highlights that are understandable on their own to avoid confusion.
Our method combines determinantal point processes and deep contextualized representations to identify an optimal set of sub-sentence segments that are both important and non-redundant to form summary highlights. 
To demonstrate the flexibility and modeling power of our method, we conduct extensive experiments on summarization datasets. 
Our analysis provides evidence that highlighting is a promising avenue of research towards future summarization.

\end{abstract}

\section{Introduction}
\label{sec:intro}

A summary is reliable only if it is true-to-original.
Abstractive summarizers are considered to be less reliable despite their impressive performance on benchmark datasets, because they can hallucinate facts and struggle to keep the original meanings intact~\cite{kryscinski-etal-2019-neural,lebanoff-etal-2019-analyzing}.
In this paper, we seek to generate summary highlights to be overlaid on the original documents to allow summaries to be understood in context and avoid misdirecting readers to false conclusions.
This is especially important in areas involving legislation, political speeches, public policies, social media, and more~\cite{Sadeh:2013,kornilova-eidelman-2019-billsum}.
Highlighting is most commonly used in education to make important information stand out and bring attention of readers to the essential topics~\cite{rello-etal-2014-keyword}.

\begin{table}[t]
\setlength{\tabcolsep}{0pt}
\renewcommand{\arraystretch}{1.15}
\centering
\begin{scriptsize}
\textsf{
\begin{tabular}{l}
\textbf{Original Document and Summary Highlights}\\[0.2em]
\toprule
{\cellcolor[gray]{.97}}Afghan opium kills 100,000 people every year worldwide -- more\\
{\cellcolor[gray]{.97}}than any other drug -- and the opiate heroin kills five times as many\\
{\cellcolor[gray]{.97}}people in NATO countries each year than the eight-year total of NATO\\
{\cellcolor[gray]{.97}}troops killed in Afghan combat, \emph{\colorbox{yellow!70}{the United Nations said Wednesday.}}\\ 
{\cellcolor[gray]{.97}}About 15 million people around the world use heroin, opium or\\ 
{\cellcolor[gray]{.97}}morphine, fueling a \$65 billion market for the drug and also fueling\\
{\cellcolor[gray]{.97}}terrorism and insurgencies...\emph{\colorbox{yellow!70}{Drug money is funding insurgencies}}\\
{\cellcolor[gray]{.97}}in Central Asia, which has huge energy reserves, Costa said...\\
{\cellcolor[gray]{.97}}Europe and Russia together consume just under half of the heroin\\
{\cellcolor[gray]{.97}}coming out of Afghanistan, the United Nations concluded, and\\
{\cellcolor[gray]{.97}}\emph{\colorbox{yellow!70}{Iran is by far the single largest consumer of Afghan opium.}}\\
\end{tabular}}
\end{scriptsize}
\caption{An example of sub-sentence highlights overlaid on the original document; the highlights are self-contained.}
\label{tab:example}
\vspace{-0.15in}
\end{table}

The characteristics of summary highlights are:
\emph{saliency}, i.e., highlights must give the main points of the documents, 
and \emph{non-redundancy}, suggesting that redundant content should not appear in a summary~\cite{Nenkova:2011}.
Importantly, a highlighted text should be \emph{self-contained}, i.e., understandable on its own, without the need for specific information from surrounding context.
Table~\ref{tab:example} provides an example of sub-sentence highlights.
In contrast, ``\emph{New Jersey is located in}'' hardly constitutes a good highlight because the information it contains is incomplete and may confuse readers.
To date, there has not been any unified framework to account for all these characteristics to generate highlights.
We overcome the challenge by identifying self-contained sub-sentence segments from the documents, then combining determinantal point processes and deep contextualized representations to produce highlights.

Determinantal point process belongs to a class of optimization methods that have had considerable success in summarizing text and video~\cite{Kulesza:2012,Gong:2014,Sharghi:2018}.
It selects a diverse subset from a ground set of items, where an item is a candidate text segment in the context of generating summary highlights.
An item is characterized by a \emph{quality} score that indicates the salience of the segment and a \emph{diversity} score that models \emph{pairwise repulsion}, suggesting that two segments carrying similar meaning cannot both be included in the summary to avoid redundancy.
The quality and diversity decomposition of DPP allows it to identify an optimal subset from a collection of candidate segments.

We study sub-sentence segments as they strike a balance between the quality and amount of highlights.
Whole sentences often contain excessive or unwanted details; keywords are succinct but less informative.
We conjecture that sub-sentence segments can be identified from a document similar to salient objects are identified from an image using bounding boxes~\cite{Girshick:2014:RFH:2679600.2679851}.
To best estimate the size of segments, we present a novel method to ``overgenerate'' a rich set of self-contained, partially-overlapping sub-sentence segments from any sentence based on contextualized representations~\cite{NIPS2019_8812,devlin-etal-2019-bert}, then leverage determinantal point processes to identify an essential subset based on saliency and non-redundancy criteria. 
Our contributions of this work are summarized as follows.

\begin{itemize}[topsep=5pt,itemsep=0pt,leftmargin=*]
\item We propose to generate sub-sentence summary highlights to be overlaid on source documents to enable users to quickly navigate through content. Comparing to keywords or whole sentences, sub-sentence segments allow us to attain a good balance between quality and amount of highlights.

\item Importantly, sub-sentence segments are designed to be self-contained, and for which we introduce a new algorithm based on deep contextual representations to obtain self-contained text segments.
All candidate segments are fed to determinantal point processes to identify an optimal subset containing informative, non-redundant, and self-contained sub-sentence highlights.

\item We perform experiments on benchmark summarization datasets to demonstrate the flexibility and modeling power of our approach.
Our analysis provides further evidence that highlighting offers a promising avenue of research.\footnote{Our source code is publicly available at \url{https://github.com/ucfnlp/better-highlighting}}

\end{itemize}

\begin{table}[t]
\setlength{\tabcolsep}{0pt}
\renewcommand{\arraystretch}{1.1}
\centering
\begin{scriptsize}
\textsf{
\begin{tabular}{l}
\textbf{Original Sentence}\\
\toprule
$\bullet$ Some interstates are closed and hundreds of flights have been \\
\,\,\,\, canceled as winter storms hit during one of the year's busiest\\
\,\,\,\, travel weeks.\\[0.8em]
\textbf{Self-Contained Segments}\\
\toprule
$\bullet$ Some interstates are closed\\
$\bullet$ hundreds of flights have been canceled as winter storms hit\\
$\bullet$ flights have been canceled as winter storms hit\\
$\bullet$ winter storms hit during one of the year’s busiest travel weeks\\[0.8em]
\textbf{Non-Self-Contained Segments}\\
\toprule
$\bullet$ Some interstates are\\
$\bullet$ closed and hundreds of flights have been\\
$\bullet$ been canceled as winter storms hit during one of\\
$\bullet$ hit during one of the year's\\
\end{tabular}}
\end{scriptsize}
\caption{Examples of self-contained and non-self-contained segments extracted from a document sentence.}
\label{tab:self-contained}
\vspace{-0.15in}
\end{table}

\section{Related Work}
\label{sec:related}

An abstract failing to retain the original meaning poses a substantial risk of harm to applications.
Abstractive summarizers can copy words from source documents or generate new words~\cite{see-etal-2017-get,tan-etal-2017-abstractive,chen-bansal-2018-fast,narayan-etal-2018-dont,gehrmann-etal-2018-bottom,liu-lapata-2019-hierarchical,laban-etal-2020-summary}.
With greater flexibility comes increased risk. 
Failing to accurately convey the original meaning can hinder the deployment of summarization techniques in real-world scenarios, as inaccurate and untruthful summaries can lead the readers to false conclusions~\cite{Cao:2018,falke-etal-2019-ranking,lebanoff-etal-2019-analyzing}.
We aim to produce summary highlights in this paper, which will be overlaid on source documents to allow summaries to be interpreted in context.

Generation of summary highlights is of crucial importance to tasks such as producing informative snippets from search outputs~\cite{kaisser-etal-2008-improving}, summarizing viewpoints in opinionated text~\cite{paul-etal-2010-summarizing,amplayo-lapata-2020-unsupervised}, and annotating website privacy policies to assist users in answering important questions~\cite{Sadeh:2013}. 
Determining the most appropriate textual unit for highlighting, however, has been an understudied problem.
Extractive summarization selects whole sentences from documents; a sentence can contain 20 to 30 words on average~\cite{kamigaito-etal-2018-higher}.
Keyphrases containing two to three words are much less informative~\cite{hasan-ng-2014-automatic}.
Neither are ideal solutions. There is a rising need for other forms of highlighting, and 
we explore sub-sentence highlights that strike a balance between the amount and quality of emphasized content.

It is best for highlighted segments to remain self-contained.
In fact, multiple partially-overlapping and self-contained segments can exist in a sentence, as illustrated in Table~\ref{tab:self-contained}.
Identifying self-contained segments has not been thoroughly investigated in previous studies.
Woodsend and Lapata~\shortcite{woodsend-lapata-2010-automatic} propose to generate story highlights by selecting and combining phrases.
Li et al.~\shortcite{li-etal-2016-role} explore elementary discourse units generated using an RST parser as selection units.
Spala et al.~\shortcite{spala-etal-2018-web} present a crowdsourcing method for workers to highlight sentences and compare systems.
Arumae et al.~\shortcite{arumae-etal-2019-towards} propose to align human abstracts and source articles to create ground-truth highlight annotations.
Importantly, and distinguishing our work from earlier literature, we make a first attempt to generate self-contained highlights, drawing on the successes of deep contextualized representations and their extraordinary ability of encoding syntactic structure~\cite{clark-etal-2019-bert,hewitt-manning-2019-structural}.
We next discuss our method in greater detail.

\section{Our Method}
\label{sec:our_method}

We present a new method to identify self-contained segments, then select important and non-redundant segments to form a summary, as text fragments containing incomplete and disorganized information are hardly successful summary highlights.

\subsection{Self-Contained Segments}
\label{sec:self-contained}

A self-contained segment is, in a sense, a miniature sentence. 
Any text segment containing incomplete or ungrammatical constructions is incomprehensible to humans.
Table~\ref{tab:self-contained} presents examples of self-contained and non-self-contained segments.
Since its very inception~\cite{vladutz-1983-natural}, the concept of ``semantically self-contained segment'' has not been sufficiently examined in the literature and lacks an universal definition.
We assume in this paper that a self-contained segment shall conform to certain syntactic validity constraints and there exists only weak dependencies between words that belong to the segment and those do not.

The automatic identification of self-contained segments requires more than segmentation or parsing sentences into tree structures~\cite{dozat-manning-2018-simpler}.
Self-contained segments do not necessarily correspond to constituents of the tree and further, there is no guarantee that tree constituents are self-contained.
In this paper, we define a segment to be a consecutive sequence of words, excluding segments formed by concatenating non-adjacent words from consideration.
We perform exhaustive search to analyze every segment of a given sentence to determine if it is self-contained or not.

\begin{figure}[t]
\centering
\includegraphics[width=3in]{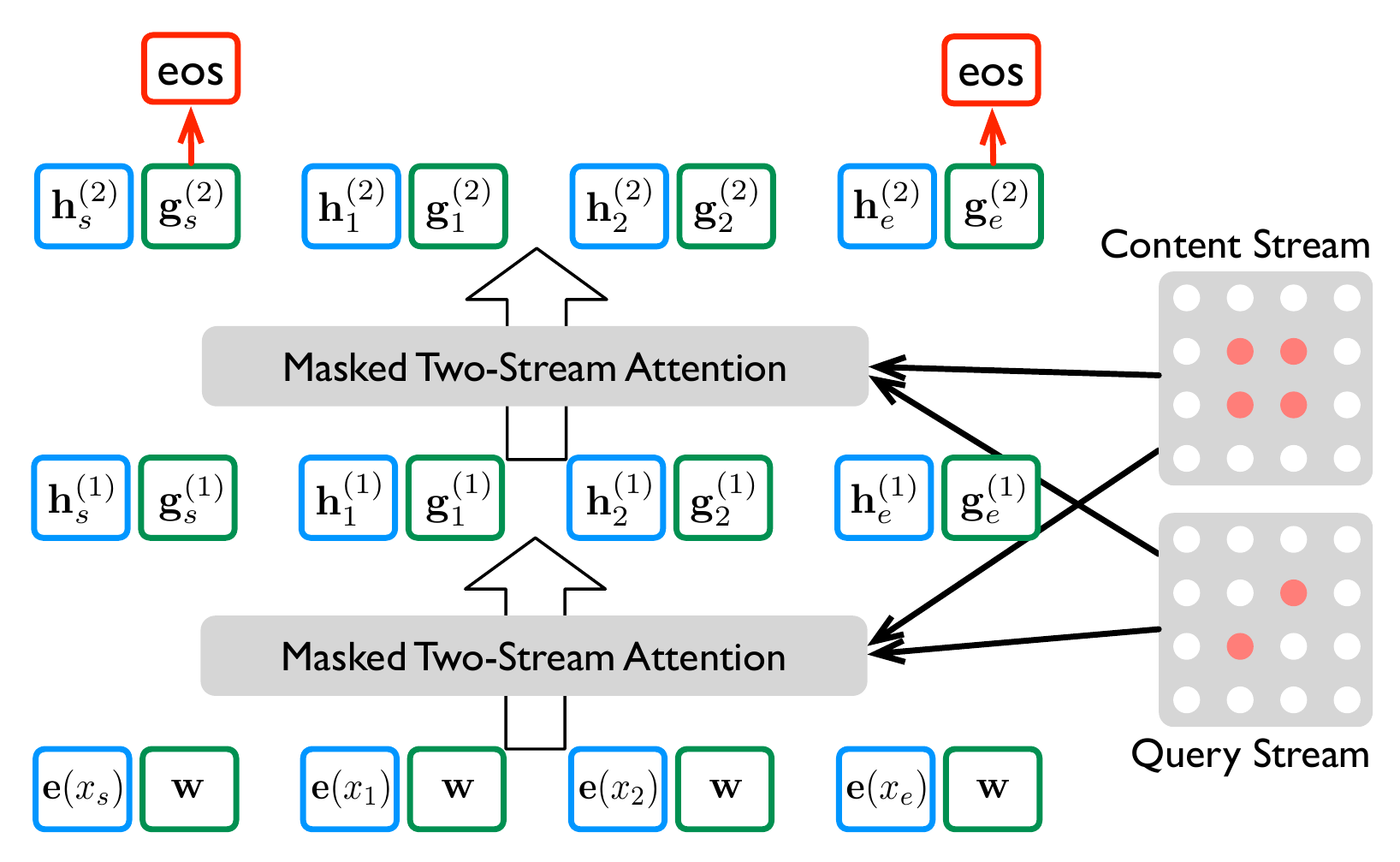}
\caption{The XLNet architecture with two-stream attention mechanism is leveraged to estimate whether a segment is self-contained or not. A self-contained segment is assumed to be preceded and followed by end-of-sentence markers (\textsf{eos}).}
\label{fig:xlnet}
\vspace{-0.1in}
\end{figure}

\begin{figure*}
\centering
\includegraphics[width=6in]{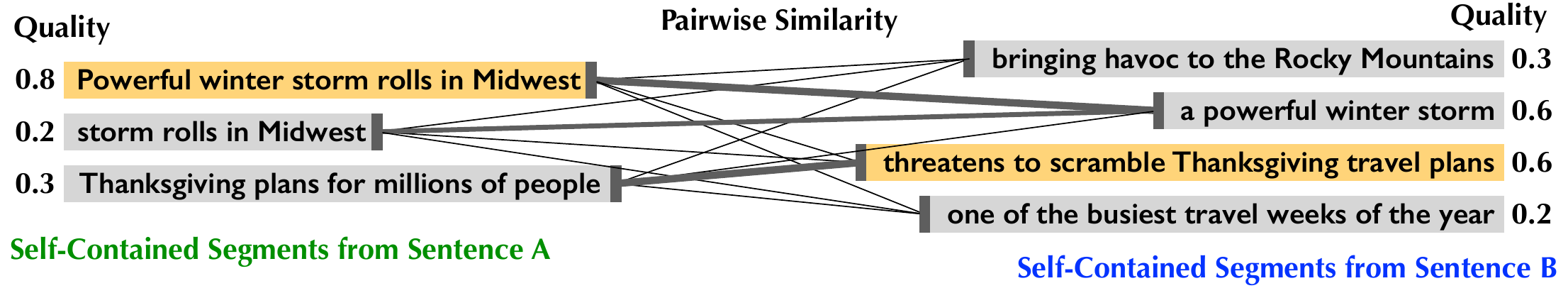}
\caption{DPP selects a set of summary segments (marked yellow) based on the \emph{quality} and \emph{pairwise dissimilarity} of segments.}
\label{fig:dpp}
\vspace{-0.1in}
\end{figure*}

Let $\mathbf{x} = [x_1, \ldots, x_{\textsf{N}}]$ be a document sentence. 
We present a method to estimate whether an arbitrary segment $\mathbf{x}_{i:j}$ of the sentence is semantically self-contained or not.
Our method is inspired by XLNet~\cite{NIPS2019_8812} that introduces a novel architecture with two-stream attention mechanism for autoregressive language modeling.
Pretrained contextualized representations such as BERT and XLNet have demonstrated remarkable success on language understanding tasks.
We expect the representations to encode the syntactic validity of segments, as similar findings are seen in recent structural probings~\cite{hewitt-manning-2019-structural}.

We hypothesize that a self-contained segment, similar to a miniature sentence, can be preceded and followed by end-of-sentence (\textsf{eos}) markers without sacrificing grammatical correctness.
We follow the convention of Clark et al.~\shortcite{clark-etal-2019-bert} to define end-of-sentence markers (\textsf{eos}) to include periods and commas.
Our method inserts hypothetical tokens $x_{\mbox{\scriptsize s}}$ and $x_{\mbox{\scriptsize e}}$ to the beginning and end positions of a segment $\mathbf{x}_{i:j}$, then constructs contextualized representations for these positions, denoted by $\mathbf{g}(\mathbf{x}_{i:j}, p_{\mbox{\scriptsize start}})$ and $\mathbf{g}(\mathbf{x}_{i:j}, p_{\mbox{\scriptsize end}})$, based on which we estimate how likely $x_{\mbox{\scriptsize s}}$ is an end-of-sentence marker $p(x_{\mbox{\scriptsize s}}\mbox{\footnotesize\textsf{=eos}} | \mathbf{x}_{i:j})$, similarly for $p(x_{\mbox{\scriptsize e}}\mbox{\footnotesize\textsf{=eos}} | \mathbf{x}_{i:j})$.
Their average probability indicates self-containedness.
A higher score of $p(z|\mathbf{x}_{i:j})$ suggests $\mathbf{x}_{i:j}$ has a higher likelihood of being self-contained.
{\medmuskip=1mu
\thinmuskip=1mu
\thickmuskip=1mu
\nulldelimiterspace=2pt
\scriptspace=1pt
\begin{align*}
p(z|\mathbf{x}_{i:j}) = \frac{1}{2} \Big( p(x_{\mbox{\scriptsize s}}\mbox{\footnotesize\textsf{=eos}} | \mathbf{x}_{i:j}) + p(x_{\mbox{\scriptsize e}}\mbox{\footnotesize\textsf{=eos}} | \mathbf{x}_{i:j})\Big)
\\
p(x_{\mbox{\scriptsize s}}\mbox{\footnotesize\textsf{=eos}} | \mathbf{x}_{i:j}) = \frac{\exp(\mathbf{e}(x_{\mbox{\scriptsize s}})^\top \mathbf{g}(\mathbf{x}_{i:j}, p_{\mbox{\scriptsize start}}))}{\sum_{x'} \exp(\mathbf{e}(x')^\top \mathbf{g}(\mathbf{x}_{i:j}, p_{\mbox{\scriptsize start}}))}
\end{align*}}

It is important to induce contextualized representations for the augmented segment \emph{without} using the content of hypothetical tokens $x_{\mbox{\scriptsize s}}$ and $x_{\mbox{\scriptsize e}}$.
We leverage XLNet with two-stream attention mechanism for this purpose, as illustrated in Figure~\ref{fig:xlnet}.
For the $k$-th position ($k$=\{i:j, start, end\}) of the $l$-th layer, a \emph{content} stream builds representation $\mathbf{h}_{k}^{(l)}$ by attending to all tokens of the segment, whereas a \emph{query} stream builds representation $\mathbf{g}_k^{(l-1)}$ simultaneously without incorporating the content of the current token $x_k$, following the equations given below.
Our method builds on the pretrained XLNet model without fine-tuning. 
It relies on two-stream attention to construct deep contextualized representations $\mathbf{g}(\mathbf{x}_{i:j}, p_{\mbox{\scriptsize start}})$ and $\mathbf{g}(\mathbf{x}_{i:j}, p_{\mbox{\scriptsize end}})$, respectively for the beginning and end positions.

\vspace{0.1in}
$\mathbf{h}_{k}^{(l)}$ = Attention(Q = $\mathbf{h}_{k}^{(l-1)}$, KV = $\mathbf{h}_{i:j}^{(l-1)}$)\\
\indent $\mathbf{g}_k^{(l)}$ = Attention(Q = $\mathbf{g}_k^{(l-1)}$, KV = $\mathbf{h}_{i:j \setminus k}^{(l-1)}$)
\vspace{0.1in}

Our method is the first attempt to extract semantically \emph{self-contained} segments from whole sentences.
Segments that do not resemble ``miniature sentences'' will be given low probabilities by the method. 
E.g., ``\emph{closed and hundreds of flights have been}'' is scored low,
not only because an end-of-sentence marker rarely occurs after ``have been,'' but also the syntactic structure of the segment does not resemble that of a well-formed sentence. 

We split a sentence at punctuation and extract a number of segments from each sentence chunk.
A segment is discarded if its start (or end) probability is lower than the upper quartile value, indicating an inappropriate start (or end) point.
The remaining segments are ordered according to the average probability.
This process produces a collection of self-contained and partially-overlapping segments from a set of documents.
Next, we assess the informativeness of the segments and leverage DPP to identify a subset to form the summary highlights.

\subsection{Segment Selection with DPP}
\label{sec:dpp}

We employ the modeling framework proposed by Cho et al.~\shortcite{cho-etal-2019-improving} for modeling determinantal point processes.
DPP~\cite{Kulesza:2012} defines a probability measure $\mathcal{P}$ over all subsets ($2^{|\mathcal{Y}|}$) of a ground set containing a collection of \textsf{N} segments $\mathcal{Y} = \{1,2,\cdots,\textsf{N}\}$. 
The probability of an extractive summary, containing a subset of the segments $Y \subseteq \mathcal{Y}$, is defined by Eq.~(\ref{eq:p_y}), where 
$\mbox{det}(\cdot)$ is the determinant of a matrix;
$L \in \mathbb{R}^{\textsf{N} \times \textsf{N}}$ is a positive semi-definite matrix and
$L_{ij}$ indicates the correlation between segments $i$ and $j$;
$L_Y$ is a submatrix of $L$ containing only entries indexed by elements in $Y$;
$I$ is the identity matrix.
This definition suggests that the probability of a summary $\mathcal{P}(Y;L)$ is proportional to the determinant of $L_Y$.
\begin{align*}
&\mathcal{P}(Y;L) = \frac{\mbox{det}(L_Y)}{\mbox{det}(L + I)}, 
\numberthis\label{eq:p_y}\\
&\mathcal{L}(\boldsymbol\theta) = \sum_{i=1}^N \log \mathcal{P}(\hat{Y}^{(i)}; L^{(i)}(\boldsymbol{\theta}))
\numberthis\label{eq:loss}
\end{align*}

A decomposition exists for the $L$-ensemble matrix: 
$L_{ij} = q_i \cdot S_{ij} \cdot q_j$
where 
$q_i \in \mathbb{R}^+$ is a {quality} score of the $i$-th segment and $S_{ij}$ is a pairwise similarity score between segments $i$ and $j$.
If $q$ and $S$ are available, $\mathcal{P}(Y)$ can be computed using Eq.~(\ref{eq:p_y}).
Estimating the pairwise similarity $S$ is trivial, we refer the reader to~\cite{cho-etal-2019-multi} for details.
In this paper, we present a \emph{inverted pyramid} method to estimate the quality of segments $q$.
The quality model is parameterized by $\boldsymbol{\theta}$, thus the L-ensemble is parameterized the same, denoted by $ L^{(i)}(\boldsymbol{\theta})$ for the $i$-th instance of the dataset.
$\hat{Y}^{(i)}$ represents the ground-truth summary (Eq.~(\ref{eq:loss})). 
The model is optimized by maximizing the log-likelihood, where parameters $\boldsymbol{\theta}$ are learned during training.
As illustrated in Figure~\ref{fig:dpp}, DPP allows us to identify a set of salient and non-redundant summary segments.

\vspace{0.08in}
\noindent\textbf{Inverted pyramid}\,\,\,\,
We describe a classifier to predict if a segment of text is summary-worthy or not according to the \emph{inverted pyramid} principle.\footnote{\url{https://en.wikipedia.org/wiki/Inverted_pyramid_(journalism)}}
It is a way of front loading a story so that the reader can get the most important information first. 
E.g., the most newsworthy information such as who, what, when, where, etc. heads the article, followed by important details, and finally other general and background information.
The inverted pyramid explains the common observation that lead baselines consisting of the first few sentences of an article perform strongly in the news domain.

Our classifier assigns a high score to a segment if its content is relevant to the lead paragraph, and a low score if its content overlaps with the bottom paragraph of a news article, which usually contains trivial details.
Importantly, the classifier is trained using CNN/DM~\cite{see-etal-2017-get}, rather than any multi-document summarization data.

During training, we obtain the ground-truth summary of each article.
A summary sentence is paired with the lead paragraph of the article that contains the top-5 sentences to form a \emph{positive} instance and similarly, with bottom-5 sentences to form a \emph{negative} instance.
If a summary sentence appears as-is in the top or bottom paragraph, we exclude the sentence from the paragraph to avoid overfitting the classifier.
At test time, the classifier learns to distill the essential content of the segment and assigns a high score to it if its content is similar to the lead paragraph, indicating the segment is relevant and summary-worthy.

For each instance, we obtain deep contextualized representation for it using the BERT architecture, where a segment and a lead (or bottom) paragraph is used as the input and the top layer hidden vector of the {\footnotesize\textsf{[CLS]}} token is extracted as the representation.
It is fed to a feedforward, a dropout and a softmax layer to predict a binary label for the segment. 
Once the model is trained, we apply it to a segment and its lead paragraph to produce a vector which is used as part of the features for computing $q$.

\vspace{0.08in}
\noindent\textbf{DPP training.}\quad
We obtain feature representations for the $i$-th segment by concatenating the previous vector and a number of surface features extracted for segment $i$. 
The features include the length and position of the segment within a document, the cosine similarity between the segment and document TF-IDF vectors~\cite{Kulesza:2011}. We abstain from using sophisticated features to avoid model overfitting.
The feature parameters $\boldsymbol{\theta}$ are to be learned during DPP training.

DPP is trained on multi-document summarization data by maximizing log-likelihood.
At each iteration, we project the L-ensemble onto the positive semi-definite (PSD) cone to ensure that it satisfies the PSD property (\S\ref{sec:dpp}). 
This is accomplished in two steps, where $L'$ is the new L-ensemble.

\vspace{0.05in}
\noindent$L = \sum_{i=0}^n \lambda_i v_i v_i^\top \,\, \textnormal{(Eigenvalue decomposition)}$\\[0.6em]
\noindent$L' = \sum_{i=0}^n \max\{\lambda_i,0\} v_i v_i^\top \,\,\,\, \textnormal{(PSD projection)}$\\[0.6em]

\begin{table}[t]
\setlength{\tabcolsep}{4.6pt}
\renewcommand{\arraystretch}{1.2}
\centering
\begin{small}
\begin{tabular}{lrrr}
\textbf{DUC-04 Test Set} & \textbf{R-1} & \textbf{R-2} & \textbf{R-SU4} \\
\toprule
DPP-BERT{\scriptsize~\cite{cho-etal-2019-multi}}  & 39.05 & 10.23 & 14.35 \\
DPP{\scriptsize~\cite{Kulesza:2012}} & 38.10 & 9.14 & 13.40 \\
SumBasic{\scriptsize~\cite{Vanderwende:2007}} & 29.48 & 4.25 & 8.64\\
KLSumm{\scriptsize(Haghighi et al., 2009)\nocite{Haghighi:2009}} & 31.04 & 6.03 & 10.23 \\
LexRank{\scriptsize~\cite{Erkan:2004}} & 34.44 & 7.11 & 11.19 \\
Centroid{\scriptsize~\cite{Hong:2014}} & 35.49 & 7.80 & 12.02 \\
ICSISumm{\scriptsize~\cite{Gillick:2009:NAACL}} & 37.31 & 9.36 & 13.12 \\
Opinosis{\scriptsize~\cite{Ganesan:2010}} & 27.07 & 5.03 & 8.63\\
Pointer-Gen{\scriptsize~\cite{see-etal-2017-get}} & 31.43 & 6.03 & 10.01\\
CopyTrans{\scriptsize~\cite{gehrmann-etal-2018-bottom}} & 28.54 & 6.38 & 7.22\\
Hi-MAP{\scriptsize~\cite{fabbri-etal-2019-multi}} & 35.78 & 8.90 & 11.43\\
\midrule
HL-TreeSegs (Our work) & {39.18} & {10.30} & {14.37} \\
HL-XLNetSegs (Our work) & \textbf{39.26} & \textbf{10.70} & \textbf{14.47} \\
\bottomrule
\end{tabular}
\end{small}
\caption{Results on DUC-04 dataset evaluated by ROUGE. 
}
\label{tab:results_duc04}
\vspace{-0.1in}
\end{table}

\begin{table*}
\setlength{\tabcolsep}{5pt}
\renewcommand{\arraystretch}{1}
\begin{scriptsize}
\begin{minipage}{\textwidth}
\centering
\textsf{
\begin{tabular}[t]{|p{6in}|}
\hline
\textbf{Human Abstract}\Tstrut\\[1mm]
$\bullet$ Exxon and Mobil discuss combining business operations.\\[1mm]
$\bullet$ A possible Exxon-Mobil merger would reunite 2 parts of Standard Oil broken up by the Supreme Court in 1911.\\[1mm]
$\bullet$ Low crude oil prices and the high cost of exploration are motives for a merger that would create the world's largest oil company.\\[1mm]
$\bullet$ As Exxon-Mobil merger talks continue, stocks of both companies surge.\\[1mm]
$\bullet$ The merger talks show that corporate mergers are back in vogue.\\[1mm]
$\bullet$ Antitrust lawyers, industry analysts, and government officials say a merger would require divestitures.\\[1mm]
$\bullet$ A Mobil employee worries that a merger would put thousands out of work, but notes that his company's stock would go up.\\[1mm]
\hline
\hline
\textbf{Highlighting (Tree Segments)}\Tstrut\\[1mm]
$\bullet$ \textcolor{dodgerblue}{Whether or not the talks between Exxon and Mobil lead to a merger or some other business combination}, America's economic history is already being rewritten.\\[1mm]
$\bullet$ \textcolor{dodgerblue}{The boards of Exxon Corp. and Mobil Corp. are expected to meet Tuesday to consider a possible merger agreement that would form the world's largest oil company}, a source close to the negotiations said Friday. \\[1mm]
$\bullet$ Exxon Corp. and Mobil Corp. \textcolor{dodgerblue}{have held discussions about combining their business operations, a person involved in the talks said Wednesday}.\\[1mm]
$\bullet$ News \textcolor{dodgerblue}{that Exxon and Mobil, two giants in the energy patch, were in merger talks last week} is the biggest sign yet that corporate marriages are back in vogue. (Rest omitted.)\\[1mm] 
\hline
\hline
\textbf{Highlighting (XLNet Segments)}\Tstrut\\[1mm]
$\bullet$ \textcolor{fuchsia}{Whether or not the talks between Exxon and Mobil lead to a merger} or some other business combination, America's economic history is already being rewritten.\\[1mm]
$\bullet$ Still, it boggles the mind to accept \textcolor{fuchsia}{the notion that hardship is driving profitable Big Oil to either merge}, as British Petroleum and Amoco have already agreed to do, or at least to consider the prospect, as Exxon and Mobil are doing.\\[1mm]
$\bullet$ Oil stocks led the way as investors soaked up the news of continuing talks between \textcolor{fuchsia}{Exxon and Mobil on a merger that would create the world's largest oil company}.\\[1mm]
$\bullet$ Although the companies only confirmed that they were discussing the possibility of a merger, \textcolor{fuchsia}{a person close to the discussions said the boards of both Exxon and Mobil} were expected to meet Tuesday to consider an agreement.\\[1mm]
$\bullet$ Analysts predicted \textcolor{fuchsia}{that there would be huge cuts in duplicate staff from both companies}, which employ 122,700 people. (Rest omitted.)\\[1mm]
\hline
\end{tabular}}
\end{minipage}
\end{scriptsize}
\caption{
Example system outputs for a topic in DUC-04. Our highlighting method is superior to sentence extraction as it allows readers to quickly skim through a large amount of text to grasp the main points. 
XLNet segments are better than tree segments. Not only can they aid reader comprehension but they are also self-contained and more concise. 
}
\label{tab:results-example}
\vspace{-0.1in}
\end{table*}

\vspace{-0.3in}
\section{Experiments}
\label{sec:experiments}

\subsection{Data Sets}

Our data comes from NIST. We use them to investigate the feasibility of the proposed multi-document summarization method.
Particularly, we use DUC-03/04~\cite{Over:2004} and TAC-08/09/10/11 datasets~\cite{Dang:2008}, which contain 60/50/48/44/46/44 document sets respectively.
These datasets are previously used as benchmarks for multi-document summarization competitions.\footnote{\url{https://tac.nist.gov/data/}\\\url{https://duc.nist.gov/data/}}
Our task is to generate a summary of less than 100 words from a set of 10 news documents, where a summary contains a set of selected text segments.
There are four human reference summaries for each document set, created by NIST evaluators. 

A system summary is evaluated against human reference summaries using ROUGE~\cite{Lin:2004}\footnote{with options \textsf{-n 2 -m -w 1.2 -c 95 -r 1000 -l 100}}, where R-1, R-2, and R-SU4 respectively measure the overlap of unigrams, bigrams and skip bigrams (with a maximum gap of 4 words) between system and reference summaries.
In the following sections, we report results on DUC-04 (trained on DUC-03) and TAC-11 (trained on TAC-08/09/10) as they are the standard test sets~\cite{Hong:2014}.

\begin{table}[t]
\setlength{\tabcolsep}{4.9pt}
\renewcommand{\arraystretch}{1.2}
\centering
\begin{small}
\begin{tabular}{lrrr}
\textbf{TAC-11 Test Set} & \textbf{R-1} & \textbf{R-2} & \textbf{R-SU4} \\
\toprule
DPP-BERT{\scriptsize~\cite{cho-etal-2019-multi}} & 38.59 & 11.06 & 14.65 \\
DPP{\scriptsize~\cite{Kulesza:2012}} & 36.95 & 9.83 & 13.57 \\
SumBasic{\scriptsize~\cite{Vanderwende:2007}} & 31.58 & 6.06 & 10.06\\
KLSumm{\scriptsize~(Haghighi et al., 2009)\nocite{Haghighi:2009}} & 31.23 & 7.07 & 10.56 \\
LexRank{\scriptsize~\cite{Erkan:2004}} & 33.10 & 7.50 & 11.13 \\
Opinosis{\scriptsize~\cite{Ganesan:2010}} & 25.15 & 5.12 & 8.12\\
Pointer-Gen{\scriptsize~\cite{see-etal-2017-get}} & 31.44 & 6.40 & 10.20\\
\midrule
HL-XLNetSegs (Our work) & {36.50} & {9.76} & {13.34} \\
HL-TreeSegs (Our work) & \textbf{37.24} & \textbf{10.04} & \textbf{13.49} \\
\bottomrule
\end{tabular}
\end{small}
\caption{ROUGE results on the TAC-11 dataset. 
}
\label{tab:results_tac11}
\vspace{-0.1in}
\end{table}

\subsection{Experimental Settings}
Our method of estimating self-containedness uses the pretrained XLNet-\textsc{large}~\cite{NIPS2019_8812} to estimate the probability of end-of-sentence markers.
We require a candidate segment to contain five or more words.
Our classifier is based on the BERT-\textsc{base} model and it is fine-tuned for two epochs on the training data. 
The maximum sequence length of the model is 512 tokens and the batch size is set to 16.
We use the Adam optimizer with an initial learning rate of $5e^{-5}$, a warm-up period of 24,400 steps, corresponding to 10\% of the training data, and linear decay after that.

\subsection{Ground-Truth Segments}
Our DPP framework is fully supervised and ground-truth summary segments are required for training the DPP.
In an ideal scenario, we would have human annotators to label the ground-truth summary segments for each document set.
It is akin to label bounding boxes for objects, which allows an object detector to be trained on millions of training examples~\cite{Girshick:2014:RFH:2679600.2679851}. 
Nonetheless, human annotation is tedious, expensive and time-consuming.
We cannot afford to have human annotators to label a large number of segments.

We introduce an approximation method instead. 
First, we greedily select a set of summary sentences from a document set that achieve the highest R-2 F-score with human reference summaries.
Secondly, for every summary sentence, we identify a single segment from a collection of over-generated and self-contained segments (\S\ref{sec:self-contained}), such that the selected attains the highest R-2 F-score with human summaries. 
Such segments are labelled as positive.
This two-step process allows for easy generation of ground-truth summary segments.

\begin{figure*}
\centering
\includegraphics[width=6.3in]{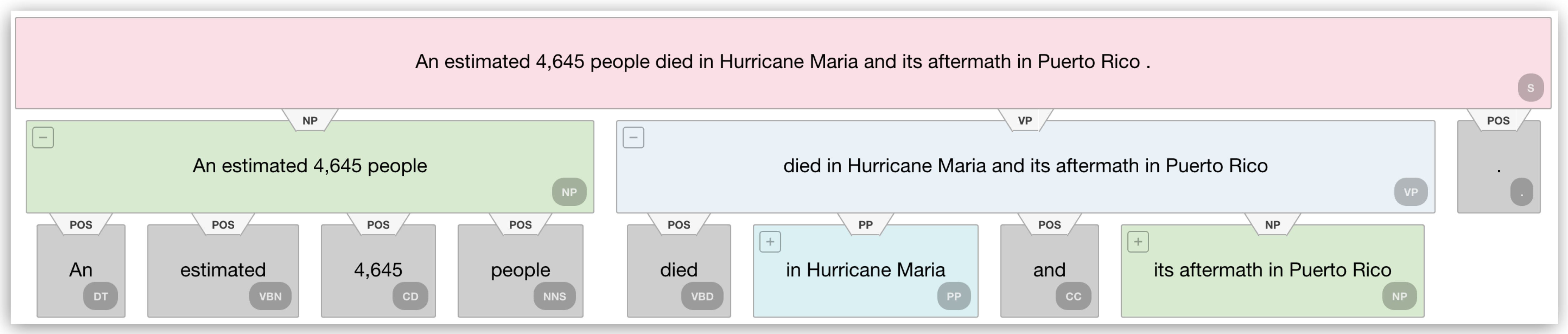}
\caption{Example of a constituent parse tree, from which tree segments are extracted.}
\label{fig:sub-tree-vis}
\vspace{-0.1in}
\end{figure*}

\subsection{Summarization Results}
\label{sec:results-summ}

We compare our method with strong extractive and abstractive summarization systems for multi-document summarization, results are shown in Tables~\ref{tab:results_duc04} and~\ref{tab:results_tac11}.
\textit{DPP}~\cite{Kulesza:2012} and variant \textit{DPP-BERT}~\cite{cho-etal-2019-multi} use determinantal point processes to extract \emph{whole sentences} from a set of documents.
\textit{SumBasic}~\cite{Vanderwende:2007} is an extractive approach leveraging the fact that frequently occurring words are more likely to be included in the summary.
\textit{KL-Sum}~\cite{Haghighi:2009} is a greedy approach that iteratively adds sentences to the summary to minimize KL divergence.
\textit{LexRank}~\cite{Erkan:2004} is a graph-based approach estimating sentence importance based on eigenvector centrality.
All of these methods extract whole sentences rather than segments from a set of documents.

We further consider abstractive summarization methods. 
\textit{Opinosis}~\cite{Ganesan:2010} creates a word co-occurrence graph and searches for a graph path to generate an abstract.
\textit{PointerGen}~\cite{see-etal-2017-get} learns to reuse source words or predict new words. 
The documents are concatenated to serve as input.
\textit{CopyTrans}~\cite{gehrmann-etal-2018-bottom} uses a 4-layer Transformer for the encoder and decoder. 
\textit{Hi-MAP}~\cite{fabbri-etal-2019-multi} introduces an end-to-end hierarchical attention model to generate abstracts from multi-document inputs.

\begin{table}[t]
\setlength{\tabcolsep}{2pt}
\renewcommand{\arraystretch}{1.1}
\centering
\begin{scriptsize}
\textsf{
\begin{tabular}{lll}
\toprule
\multicolumn{3}{l}{\textbf{Segments and Scores of Self-Containedness}}\\[0.8em]
1. & 0.646 & winter storms hit during one of the year's busiest\\
& & travel weeks\\
2. & 0.644 & storms hit during one of the year's busiest travel weeks\\
3. & 0.584 & of the year's busiest travel weeks\\
4. & 0.525 & one of the year's busiest travel weeks\\
$\ldots$ & $\ldots$ & $\ldots$\\
10. & 0.132 & and hundreds of flights have been canceled as winter\\
& & storms hit during one of the year's busiest travel weeks\\
11. & 0.122 & and hundreds of flights have been canceled\\
& & as winter storms hit\\
$\ldots$ & $\ldots$ & $\ldots$\\
150. & 0.0019 & of flights have been canceled as winter\\
151. & 0.0014 & Some interstates are closed and hundreds of flights\\
& & have been canceled as winter\\
152. & 0.0013 & hundreds of flights have been canceled as winter\\
153. & 0.0008 & are closed and hundreds of flights have been\\
& & canceled as winter\\
\bottomrule
\end{tabular}}
\end{scriptsize}
\caption{Examples of segments generated by XLNet and their scores of self-containedness.}
\label{tab:self-contained-score}
\vspace{-0.15in}
\end{table}

We explore two variants of our proposed method, called \textit{HL-XLNetSegs} and \textit{HL-TreeSegs}, focusing on highlighting summary segments.
The former utilizes XLNet to extract a set of partially-overlapping segments from a sentence; 
the latter decomposes a sentence constituent parse tree into subtrees and collect text segments governed by the subtrees. 
An illustration is shown in Figure~\ref{fig:sub-tree-vis}.
Constituent parse trees are obtained using the Stanford parser~\cite{Manning:2014}.
In both cases, the segments are passed to DPP, which identifies a set of important and non-redundant segments as highlights.

As shown in Tables~\ref{tab:results_duc04} and~\ref{tab:results_tac11}, we find both methods to perform competitively with state-of-the-art extractive and abstractive systems, while producing summary segments with simpler structure.
Our \textit{HL-XLNetSegs} method achieves the highest scores on DUC-04 and it performs comparable to other systems on TAC-11.\footnote{
Our preliminary experiment comparing the quality classifier against that of Cho et al.~\shortcite{cho-etal-2019-multi} shows that our method obtains significantly better classification accuracy (70\% vs. 96\%) when evaluated on the CNN/DM test set with balanced positive/negative examples. With a new and improved quality classifier, we expect the current results to surpass that of running the Cho et al. system with self-contained segments.}
It is important to note that breaking a sentence into smaller segments dramatically increases the search space, making it a challenging task to accurately identify summary segments, yet extracting segments remains necessary as whole sentences may contain excessive and unwanted details.
The degree of difficulty involved in generating sub-sentence highlights is thus beyond that of sentence selection.
A similar finding is reported by~\cite{cheng-lapata-2016-neural}.

\begin{table}[t]
\setlength{\tabcolsep}{6pt}
\renewcommand{\arraystretch}{1.1}
\centering
\begin{small}
\begin{tabular}{lrr}
& \textbf{DUC} & \textbf{TAC}\\
\toprule
\# Words per XLNet segment & 9.55 & 8.05\\
\# XLNet segments per sentence & 2.48 & 2.49\\
\# Total segments per document set & 398 & 352\\
\# Summary segments per document set & 9.62 & 9.09\\
\midrule
\# Words per tree segment & 12.89 & 13.94\\
\# Tree segments per sentence & 3.31 & 3.33\\
\# Total segments per document set & 549 & 478\\
\# Summary segments per document set & 13.68 & 16.56\\
\bottomrule
\end{tabular}
\end{small}
\caption{Statistics of text segments generated by XLNet and the constituent parse tree method on DUC/TAC datasets.}
\label{tab:stats-self-contained}
\vspace{-0.15in}
\end{table}

Table~\ref{tab:stats-self-contained} presents a direct comparison of XLNet and tree segments on DUC and TAC datasets. 
We find that XLNet segments are more concise than tree segments.
A tree segment contains 13 tokens on average, while an XLNet segment contains 9.6 tokens on DUC-04.
Both methods produce a large number of candidate segments, ranging from 350 to 550 segments per document set, with only 9 to 17 ground-truth summary segments per document set.
The small ratio poses a substantial challenge for DPP.
Not only must it identify salient content but it has to accurately identify the segments worthy of being included in the summary.
In Table~\ref{tab:results-example}, we show example highlights produced by both methods; more examples are in supplementary.

\begin{figure}[t]	
	\centering
	\includegraphics[width=3in]{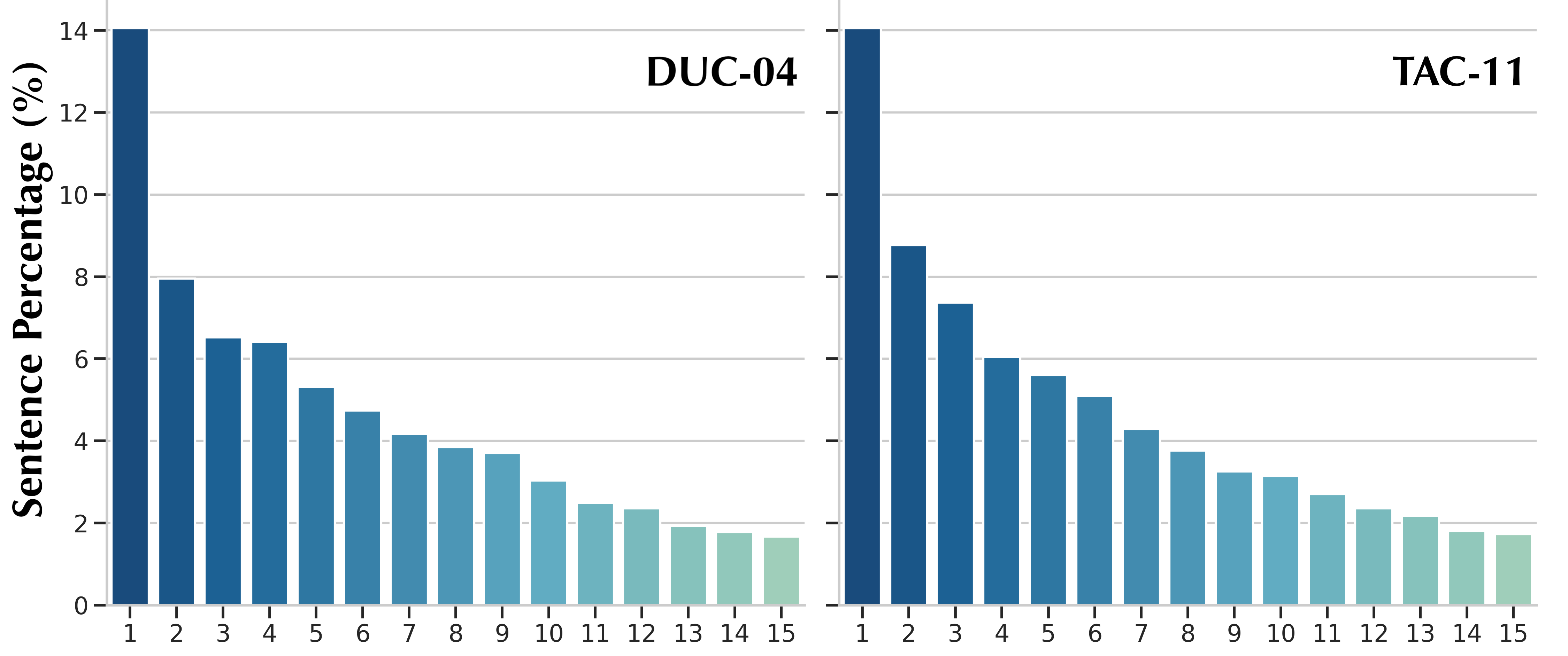}
	\vspace{-0.25in}
	\label{fig:full_sent_pos}
	\caption{Absolute position of the whole sentence among all segments sorted by XLNet scores of self-containedness.}
	\label{fig:orig_sent_pos}
\vspace{-0.1in}
\end{figure}

Segments generated by XLNet are sorted according to their scores of self-containedness, $p(z|\mathbf{x}_{i:j})$. 
In Table~\ref{tab:self-contained-score}, we provide examples of segments and their scores.
The higher the score, the more likely the segment resembles a ``miniature sentence.''
We are particularly interested in understanding where the original sentence is placed according to XLNet scores; results are shown in Figure~\ref{fig:orig_sent_pos}.
We observe that in 60\% of the cases, the original sentence is placed among the top-10 candidates, suggesting the effectiveness of the XLNet model.
As segments are shorter and occur more often in natural language texts, it is possible that they are considered more self-contained than the original sentence.

Segments extracted from subtrees are sorted by the depth of tree nodes.
The higher nodes are informative constituents denoting complex noun phrases and sentential clauses~\cite{hwa-1999-supervised}. 
An important caveat of the tree segments is their lack of \emph{coverage}.
E.g., ``\emph{4,645 people died}'' is a valid self-contained segment, but it does not belong to a tree constituent, as seen in Figure~\ref{fig:sub-tree-vis}.
Given that drawback, we focus on segments created by XLNet in our experiments.\footnote{Extracting propositions from a given sentence is beyond the scope of this paper, as proposition structures given by OpenIE~\cite{Etzioni2007} or PropS~\cite{stanovsky2016getting} are often not consecutive segments of text. Instead, they are presented as relation tuples or directed graphs. Highlighting the proposition structures can cause hundreds of small text chunks to be highlighted in the documents, which may result in undesirable visual effects.}

\begin{table}[t]
\setlength{\tabcolsep}{4.5pt}
\renewcommand{\arraystretch}{1.1}
\centering
\begin{small}
\begin{tabular}{lcccc}
& \multicolumn{4}{c}{\textbf{Self-Containedness Score}}\\ 
XLNet & $\geq$3(\%) & $\geq$4(\%) & =5(\%) & Average\\
\toprule
All Segments & 54.86 & 30.00 & 10.68 & 2.80\\
Top-5 Segments & 55.25 & 30.24 & 10.78 & 2.81\\
Top-3 Segments & \textbf{61.04} & \textbf{34.04} & \textbf{12.42} & \textbf{2.95}\\
\bottomrule
\end{tabular}
\end{small}
\caption{
Human evaluation of the self-containedness of text segments. 
The top-3 segments of XLNet exhibit a high degree of self-containedness:
61\% of them have an average score of 3 or above, 34\% have $\geq$4 score, and 12\% receive the full score.  
}
\label{tab:human-eval}
\end{table}

\begin{table}[t]
\setlength{\tabcolsep}{0pt}
\renewcommand{\arraystretch}{1.1}
\begin{scriptsize}
\begin{minipage}[b]{\hsize} 
\centering
\textsf{
\begin{tabular}[t]{p{3in}} 
\textbf{\textcolor{blue}{[Original Sentence]} District Attorney David Roger agreed to drop charges including kidnapping, armed robbery, assault with a deadly weapon and conspiracy against both men.}\\[1mm]
\toprule
$\bullet$ District \textcolor{orangered}{Attorney David Roger agreed to drop charges including kidnapping}, armed robbery, assault with a deadly weapon and conspiracy against both men. (4.0)\\[1mm]
$\bullet$ District \textcolor{darkgreen}{Attorney David Roger agreed to drop charges} including kidnapping, armed robbery, assault with a deadly weapon and conspiracy against both men. (3.8)\\[1mm]
$\bullet$ District Attorney David Roger agreed to drop charges including kidnapping, armed robbery, \textcolor{dodgerblue}{assault with a deadly weapon and conspiracy} against both men. (3.6)\\[1mm]
\bottomrule
\end{tabular}}
\end{minipage}
\hfill
\end{scriptsize}
\caption{
Example text segments produced by the XLNet algorithm. 
Each segment is judged by five human evaluators on a scale of 1 (worst) to 5 (best) and we report their average scores. 
Human evaluation suggests that text segments generated by our model demonstrate a high degree of self-containedness.
} 
\label{tab:human-eval-examples}
\vspace{-0.1in}
\end{table}

\subsection{Self-Containedness}
\label{sec:results-self-contained}

We perform further analysis to investigate the effectiveness of our method on generating self-contained segments (\S\ref{sec:self-contained}).
It is impractical to create a gold-standard by asking human annotators to judge all available sentence segments, as the number of segments is polynomial in sentence length.
Instead, we perform post-hoc evaluation on segments generated by our XLNet algorithm, which are used as input to DPP.
We sample 20 topics from TAC-11, extract 3 sentences from each document for a total of 585 sentences and 1,792 system-generated segments.
A human annotator is given the original sentence and its segments and asked to score each segment on a Likert scale of 1 (worst) to 5 (best) for self-containedness.
A Likert scale is necessary to accommodate potentially ambiguous cases.
We employ 5 human annotators to judge each segment, their average scores are reported in Table~\ref{tab:human-eval}.

We observe that 61\% of top-3 segments have an average score of $\geq$3; 34\% have a score $\geq$4; and 12\% receive the full score.
The human annotators are able to achieve a moderate level of agreement. The standard deviation of their scores is 0.95; 44\% of the segments have their majority score agreed by three or more annotators. 
Table~\ref{tab:human-eval-examples} presents example segments and their human assessment scores (more in supplementary).
While our summary highlights have been evaluated using both standard automatic metrics for assessing the informativeness of the summary and human assessment for judging the well-formedness of individual segments, we hope to explore other methods in future work, including human evaluation of highlights for the entire document set.
The task is nontrivial. It requires a well-designed, intuitive graphical user interface for evaluators to read through all source documents and their accompanying summaries/highlights~\cite{Elhadad2006}. 
Our method constitutes the preliminary step of generating summary highlights. 
This form of summarization allows readers to grasp the main points while remaining succinct and accessible, offering a promising avenue of research.

\section{Conclusion}
\label{sec:conclusion}

We make a first attempt to create sub-sentence summary highlights that are understandable and require minimum information from the surrounding context.
Highlighting is important to help readers sift through a large amount of texts and quickly grasp the main points. 
We describe a novel methodology to generate a rich set of self-contained segments from the documents, then use determinantal point processes to identify summary highlights. 
The method can be extended to other text genres such as public policies to aid reader comprehension, which will be our future work to explore.

\section*{Acknowledgments}

We are grateful to the anonymous reviewers for their insightful feedback.
We thank Chris Wiggins for discussion on the inverted pyramid principle.
This research was supported in part by the National Science Foundation grant IIS-1909603.

\bibliography{more,fei,summ,abs_summ,anthology}

\begin{thebibliography}{53}
\expandafter\ifx\csname natexlab\endcsname\relax\def\natexlab#1{#1}\fi

\bibitem[{Amplayo and Lapata(2020)}]{amplayo-lapata-2020-unsupervised}
Reinald~Kim Amplayo and Mirella Lapata. 2020.
\newblock \href {https://doi.org/10.18653/v1/2020.acl-main.175} {Unsupervised
  opinion summarization with noising and denoising}.
\newblock In \emph{Proceedings of the 58th Annual Meeting of the Association
  for Computational Linguistics}, pages 1934--1945, Online. Association for
  Computational Linguistics.

\bibitem[{Arumae et~al.(2019)Arumae, Bhatia, and
  Liu}]{arumae-etal-2019-towards}
Kristjan Arumae, Parminder Bhatia, and Fei Liu. 2019.
\newblock \href {https://doi.org/10.18653/v1/D19-5408} {Towards annotating and
  creating summary highlights at sub-sentence level}.
\newblock In \emph{Proceedings of the 2nd Workshop on New Frontiers in
  Summarization}, pages 64--69, Hong Kong, China. Association for Computational
  Linguistics.

\bibitem[{Banko et~al.(2007)Banko, Cafarella, Soderland, Broadhead, and
  Etzioni}]{Etzioni2007}
Michele Banko, Michael~J. Cafarella, Stephen Soderland, Matt Broadhead, and
  Oren Etzioni. 2007.
\newblock Open information extraction from the web.
\newblock In \emph{Proceedings of the 20th International Joint Conference on
  Artifical Intelligence}, IJCAI'07, page 2670–2676, San Francisco, CA, USA.

\bibitem[{Cao et~al.(2018)Cao, Wei, Li, and Li}]{Cao:2018}
Ziqiang Cao, Furu Wei, Wenjie Li, and Sujian Li. 2018.
\newblock \href {https://arxiv.org/abs/1711.04434} {Faithful to the original:
  {F}act aware neural abstractive summarization}.
\newblock In \emph{Proceedings of the AAAI Conference on Artificial
  Intelligence (AAAI)}.

\bibitem[{Chen and Bansal(2018)}]{chen-bansal-2018-fast}
Yen-Chun Chen and Mohit Bansal. 2018.
\newblock \href {https://doi.org/10.18653/v1/P18-1063} {Fast abstractive
  summarization with reinforce-selected sentence rewriting}.
\newblock In \emph{Proceedings of the 56th Annual Meeting of the Association
  for Computational Linguistics (Volume 1: Long Papers)}, pages 675--686,
  Melbourne, Australia. Association for Computational Linguistics.

\bibitem[{Cheng and Lapata(2016)}]{cheng-lapata-2016-neural}
Jianpeng Cheng and Mirella Lapata. 2016.
\newblock \href {https://doi.org/10.18653/v1/P16-1046} {Neural summarization by
  extracting sentences and words}.
\newblock In \emph{Proceedings of the 54th Annual Meeting of the Association
  for Computational Linguistics (Volume 1: Long Papers)}, pages 484--494,
  Berlin, Germany. Association for Computational Linguistics.

\bibitem[{Cho et~al.(2019{\natexlab{a}})Cho, Lebanoff, Foroosh, and
  Liu}]{cho-etal-2019-improving}
Sangwoo Cho, Logan Lebanoff, Hassan Foroosh, and Fei Liu. 2019{\natexlab{a}}.
\newblock \href {https://doi.org/10.18653/v1/P19-1098} {Improving the
  similarity measure of determinantal point processes for extractive
  multi-document summarization}.
\newblock In \emph{Proceedings of the 57th Annual Meeting of the Association
  for Computational Linguistics}, pages 1027--1038, Florence, Italy.
  Association for Computational Linguistics.

\bibitem[{Cho et~al.(2019{\natexlab{b}})Cho, Li, Yu, Foroosh, and
  Liu}]{cho-etal-2019-multi}
Sangwoo Cho, Chen Li, Dong Yu, Hassan Foroosh, and Fei Liu. 2019{\natexlab{b}}.
\newblock \href {https://doi.org/10.18653/v1/D19-5412} {Multi-document
  summarization with determinantal point processes and contextualized
  representations}.
\newblock In \emph{Proceedings of the 2nd Workshop on New Frontiers in
  Summarization}, pages 98--103, Hong Kong, China. Association for
  Computational Linguistics.

\bibitem[{Clark et~al.(2019)Clark, Khandelwal, Levy, and
  Manning}]{clark-etal-2019-bert}
Kevin Clark, Urvashi Khandelwal, Omer Levy, and Christopher~D. Manning. 2019.
\newblock \href {https://doi.org/10.18653/v1/W19-4828} {What does {BERT} look
  at? an analysis of {BERT}{'}s attention}.
\newblock In \emph{Proceedings of the 2019 ACL Workshop BlackboxNLP: Analyzing
  and Interpreting Neural Networks for NLP}, pages 276--286, Florence, Italy.
  Association for Computational Linguistics.

\bibitem[{Dang and Owczarzak(2008)}]{Dang:2008}
Hoa~Trang Dang and Karolina Owczarzak. 2008.
\newblock \href
  {https://www.nist.gov/publications/overview-tac-2008-update-summarization-task}
  {Overview of the {TAC} 2008 update summarization task}.
\newblock In \emph{Proceedings of Text Analysis Conference}.

\bibitem[{Devlin et~al.(2019)Devlin, Chang, Lee, and
  Toutanova}]{devlin-etal-2019-bert}
Jacob Devlin, Ming-Wei Chang, Kenton Lee, and Kristina Toutanova. 2019.
\newblock \href {https://doi.org/10.18653/v1/N19-1423} {{BERT}: Pre-training of
  deep bidirectional transformers for language understanding}.
\newblock In \emph{Proceedings of the 2019 Conference of the North {A}merican
  Chapter of the Association for Computational Linguistics: Human Language
  Technologies, Volume 1 (Long and Short Papers)}, pages 4171--4186,
  Minneapolis, Minnesota. Association for Computational Linguistics.

\bibitem[{Dozat and Manning(2018)}]{dozat-manning-2018-simpler}
Timothy Dozat and Christopher~D. Manning. 2018.
\newblock \href {https://doi.org/10.18653/v1/P18-2077} {Simpler but more
  accurate semantic dependency parsing}.
\newblock In \emph{Proceedings of the 56th Annual Meeting of the Association
  for Computational Linguistics (Volume 2: Short Papers)}, pages 484--490,
  Melbourne, Australia. Association for Computational Linguistics.

\bibitem[{Elhadad(2006)}]{Elhadad2006}
Noemie Elhadad. 2006.
\newblock \emph{User-Sensitive Text Summarization: Application to the Medical
  Domain}.
\newblock Ph.D. thesis, USA.

\bibitem[{Erkan and Radev(2004)}]{Erkan:2004}
G\"{u}nes Erkan and Dragomir~R. Radev. 2004.
\newblock \href {https://www.aaai.org/Papers/JAIR/Vol22/JAIR-2214.pdf}
  {{LexRank}: {G}raph-based lexical centrality as salience in text
  summarization}.
\newblock \emph{Journal of Artificial Intelligence Research}.

\bibitem[{Fabbri et~al.(2019)Fabbri, Li, She, Li, and
  Radev}]{fabbri-etal-2019-multi}
Alexander Fabbri, Irene Li, Tianwei She, Suyi Li, and Dragomir Radev. 2019.
\newblock \href {https://doi.org/10.18653/v1/P19-1102} {Multi-news: A
  large-scale multi-document summarization dataset and abstractive hierarchical
  model}.
\newblock In \emph{Proceedings of the 57th Annual Meeting of the Association
  for Computational Linguistics}, pages 1074--1084, Florence, Italy.
  Association for Computational Linguistics.

\bibitem[{Falke et~al.(2019)Falke, Ribeiro, Utama, Dagan, and
  Gurevych}]{falke-etal-2019-ranking}
Tobias Falke, Leonardo F.~R. Ribeiro, Prasetya~Ajie Utama, Ido Dagan, and Iryna
  Gurevych. 2019.
\newblock \href {https://doi.org/10.18653/v1/P19-1213} {Ranking generated
  summaries by correctness: An interesting but challenging application for
  natural language inference}.
\newblock In \emph{Proceedings of the 57th Annual Meeting of the Association
  for Computational Linguistics}, pages 2214--2220, Florence, Italy.
  Association for Computational Linguistics.

\bibitem[{Ganesan et~al.(2010)Ganesan, Zhai, and Han}]{Ganesan:2010}
Kavita Ganesan, ChengXiang Zhai, and Jiawei Han. 2010.
\newblock \href {https://www.aclweb.org/anthology/C10-1039} {Opinosis: {A}
  graph-based approach to abstractive summarization of highly redundant
  opinions}.
\newblock In \emph{Proceedings of the International Conference on Computational
  Linguistics (COLING)}.

\bibitem[{Gehrmann et~al.(2018)Gehrmann, Deng, and
  Rush}]{gehrmann-etal-2018-bottom}
Sebastian Gehrmann, Yuntian Deng, and Alexander Rush. 2018.
\newblock \href {https://doi.org/10.18653/v1/D18-1443} {Bottom-up abstractive
  summarization}.
\newblock In \emph{Proceedings of the 2018 Conference on Empirical Methods in
  Natural Language Processing}, pages 4098--4109, Brussels, Belgium.
  Association for Computational Linguistics.

\bibitem[{Gillick and Favre(2009)}]{Gillick:2009:NAACL}
Dan Gillick and Benoit Favre. 2009.
\newblock \href {https://dl.acm.org/citation.cfm?id=1611640} {A scalable global
  model for summarization}.
\newblock In \emph{Proceedings of the NAACL Workshop on Integer Linear
  Programming for Natural Langauge Processing}.

\bibitem[{Girshick et~al.(2014)Girshick, Donahue, Darrell, and
  Malik}]{Girshick:2014:RFH:2679600.2679851}
Ross Girshick, Jeff Donahue, Trevor Darrell, and Jitendra Malik. 2014.
\newblock \href {https://doi.org/10.1109/CVPR.2014.81} {Rich feature
  hierarchies for accurate object detection and semantic segmentation}.
\newblock In \emph{Proceedings of the 2014 IEEE Conference on Computer Vision
  and Pattern Recognition}, CVPR '14, pages 580--587, Washington, DC, USA. IEEE
  Computer Society.

\bibitem[{Gong et~al.(2014)Gong, Chao, Grauman, and Sha}]{Gong:2014}
Boqing Gong, Wei-Lun Chao, Kristen Grauman, and Fei Sha. 2014.
\newblock \href
  {https://papers.nips.cc/paper/5413-diverse-sequential-subset-selection-for-supervised-video-summarization.pdf}
  {Diverse sequential subset selection for supervised video summarization}.
\newblock In \emph{Proceedings of Neural Information Processing Systems
  (NIPS)}.

\bibitem[{Haghighi and Vanderwende(2009)}]{Haghighi:2009}
Aria Haghighi and Lucy Vanderwende. 2009.
\newblock \href {https://www.aclweb.org/anthology/N09-1041} {Exploring content
  models for multi-document summarization}.
\newblock In \emph{Proceedings of the North American Chapter of the Association
  for Computational Linguistics (NAACL)}.

\bibitem[{Hasan and Ng(2014)}]{hasan-ng-2014-automatic}
Kazi~Saidul Hasan and Vincent Ng. 2014.
\newblock \href {https://doi.org/10.3115/v1/P14-1119} {Automatic keyphrase
  extraction: A survey of the state of the art}.
\newblock In \emph{Proceedings of the 52nd Annual Meeting of the Association
  for Computational Linguistics (Volume 1: Long Papers)}, pages 1262--1273,
  Baltimore, Maryland. Association for Computational Linguistics.

\bibitem[{Hewitt and Manning(2019)}]{hewitt-manning-2019-structural}
John Hewitt and Christopher~D. Manning. 2019.
\newblock \href {https://doi.org/10.18653/v1/N19-1419} {{A} structural probe
  for finding syntax in word representations}.
\newblock In \emph{Proceedings of the 2019 Conference of the North {A}merican
  Chapter of the Association for Computational Linguistics: Human Language
  Technologies, Volume 1 (Long and Short Papers)}, pages 4129--4138,
  Minneapolis, Minnesota. Association for Computational Linguistics.

\bibitem[{Hong et~al.(2014)Hong, Conroy, Favre, Kulesza, Lin, and
  Nenkova}]{Hong:2014}
Kai Hong, John~M Conroy, Benoit Favre, Alex Kulesza, Hui Lin, and Ani Nenkova.
  2014.
\newblock \href
  {http://www.lrec-conf.org/proceedings/lrec2014/pdf/1093_Paper.pdf} {A
  repository of state of the art and competitive baseline summaries for generic
  news summarization}.
\newblock In \emph{Proceedings of the Ninth International Conference on
  Language Resources and Evaluation (LREC)}.

\bibitem[{Hwa(1999)}]{hwa-1999-supervised}
Rebecca Hwa. 1999.
\newblock \href {https://doi.org/10.3115/1034678.1034699} {Supervised grammar
  induction using training data with limited constituent information}.
\newblock In \emph{Proceedings of the 37th Annual Meeting of the Association
  for Computational Linguistics}, pages 73--79, College Park, Maryland, USA.
  Association for Computational Linguistics.

\bibitem[{Kaisser et~al.(2008)Kaisser, Hearst, and
  Lowe}]{kaisser-etal-2008-improving}
Michael Kaisser, Marti~A. Hearst, and John~B. Lowe. 2008.
\newblock \href {https://www.aclweb.org/anthology/P08-1080} {Improving search
  results quality by customizing summary lengths}.
\newblock In \emph{Proceedings of ACL-08: HLT}, pages 701--709, Columbus, Ohio.
  Association for Computational Linguistics.

\bibitem[{Kamigaito et~al.(2018)Kamigaito, Hayashi, Hirao, and
  Nagata}]{kamigaito-etal-2018-higher}
Hidetaka Kamigaito, Katsuhiko Hayashi, Tsutomu Hirao, and Masaaki Nagata. 2018.
\newblock \href {https://doi.org/10.18653/v1/N18-1155} {Higher-order syntactic
  attention network for longer sentence compression}.
\newblock In \emph{Proceedings of the 2018 Conference of the North {A}merican
  Chapter of the Association for Computational Linguistics: Human Language
  Technologies, Volume 1 (Long Papers)}, pages 1716--1726, New Orleans,
  Louisiana. Association for Computational Linguistics.

\bibitem[{Kornilova and Eidelman(2019)}]{kornilova-eidelman-2019-billsum}
Anastassia Kornilova and Vladimir Eidelman. 2019.
\newblock \href {https://doi.org/10.18653/v1/D19-5406} {{B}ill{S}um: A corpus
  for automatic summarization of {US} legislation}.
\newblock In \emph{Proceedings of the 2nd Workshop on New Frontiers in
  Summarization}, pages 48--56, Hong Kong, China. Association for Computational
  Linguistics.

\bibitem[{Kryscinski et~al.(2019)Kryscinski, Keskar, McCann, Xiong, and
  Socher}]{kryscinski-etal-2019-neural}
Wojciech Kryscinski, Nitish~Shirish Keskar, Bryan McCann, Caiming Xiong, and
  Richard Socher. 2019.
\newblock \href {https://doi.org/10.18653/v1/D19-1051} {Neural text
  summarization: A critical evaluation}.
\newblock In \emph{Proceedings of the 2019 Conference on Empirical Methods in
  Natural Language Processing and the 9th International Joint Conference on
  Natural Language Processing (EMNLP-IJCNLP)}, pages 540--551, Hong Kong,
  China. Association for Computational Linguistics.

\bibitem[{Kulesza and Taskar(2011)}]{Kulesza:2011}
Alex Kulesza and Ben Taskar. 2011.
\newblock \href {https://dl.acm.org/citation.cfm?id=3020597} {Learning
  determinantal point processes}.
\newblock In \emph{Proceedings of the Conference on Uncertainty in Artificial
  Intelligence (UAI)}.

\bibitem[{Kulesza and Taskar(2012)}]{Kulesza:2012}
Alex Kulesza and Ben Taskar. 2012.
\newblock \href {https://arxiv.org/abs/1207.6083} {\emph{Determinantal Point
  Processes for Machine Learning}}.
\newblock Now Publishers Inc.

\bibitem[{Laban et~al.(2020)Laban, Hsi, Canny, and
  Hearst}]{laban-etal-2020-summary}
Philippe Laban, Andrew Hsi, John Canny, and Marti~A. Hearst. 2020.
\newblock \href {https://doi.org/10.18653/v1/2020.acl-main.460} {The summary
  loop: Learning to write abstractive summaries without examples}.
\newblock In \emph{Proceedings of the 58th Annual Meeting of the Association
  for Computational Linguistics}, pages 5135--5150, Online. Association for
  Computational Linguistics.

\bibitem[{Lebanoff et~al.(2019)Lebanoff, Muchovej, Dernoncourt, Kim, Kim,
  Chang, and Liu}]{lebanoff-etal-2019-analyzing}
Logan Lebanoff, John Muchovej, Franck Dernoncourt, Doo~Soon Kim, Seokhwan Kim,
  Walter Chang, and Fei Liu. 2019.
\newblock \href {https://doi.org/10.18653/v1/D19-5413} {Analyzing sentence
  fusion in abstractive summarization}.
\newblock In \emph{Proceedings of the 2nd Workshop on New Frontiers in
  Summarization}, pages 104--110, Hong Kong, China. Association for
  Computational Linguistics.

\bibitem[{Li et~al.(2016)Li, Thadani, and Stent}]{li-etal-2016-role}
Junyi~Jessy Li, Kapil Thadani, and Amanda Stent. 2016.
\newblock \href {https://doi.org/10.18653/v1/W16-3617} {The role of discourse
  units in near-extractive summarization}.
\newblock In \emph{Proceedings of the 17th Annual Meeting of the Special
  Interest Group on Discourse and Dialogue}, pages 137--147, Los Angeles.
  Association for Computational Linguistics.

\bibitem[{Lin(2004)}]{Lin:2004}
Chin-Yew Lin. 2004.
\newblock \href {https://www.aclweb.org/anthology/W04-1013} {{ROUGE}: a package
  for automatic evaluation of summaries}.
\newblock In \emph{Proceedings of ACL Workshop on Text Summarization Branches
  Out}.

\bibitem[{Liu and Lapata(2019)}]{liu-lapata-2019-hierarchical}
Yang Liu and Mirella Lapata. 2019.
\newblock \href {https://doi.org/10.18653/v1/P19-1500} {Hierarchical
  transformers for multi-document summarization}.
\newblock In \emph{Proceedings of the 57th Annual Meeting of the Association
  for Computational Linguistics}, pages 5070--5081, Florence, Italy.
  Association for Computational Linguistics.

\bibitem[{Manning et~al.(2014)Manning, Surdeanu, Bauer, Finkel, Bethard, and
  McClosky}]{Manning:2014}
Christopher~D Manning, Mihai Surdeanu, John Bauer, Jenny Finkel, Steven~J
  Bethard, and David McClosky. 2014.
\newblock The stanford {CoreNLP} natural language processing toolkit.
\newblock In \emph{Proceedings of the Association for Computational Linguistics
  (ACL) System Demonstrations}.

\bibitem[{Narayan et~al.(2018)Narayan, Cohen, and
  Lapata}]{narayan-etal-2018-dont}
Shashi Narayan, Shay~B. Cohen, and Mirella Lapata. 2018.
\newblock \href {https://doi.org/10.18653/v1/D18-1206} {Don{'}t give me the
  details, just the summary! topic-aware convolutional neural networks for
  extreme summarization}.
\newblock In \emph{Proceedings of the 2018 Conference on Empirical Methods in
  Natural Language Processing}, pages 1797--1807, Brussels, Belgium.
  Association for Computational Linguistics.

\bibitem[{Nenkova and McKeown(2011)}]{Nenkova:2011}
Ani Nenkova and Kathleen McKeown. 2011.
\newblock \href {https://www.nowpublishers.com/article/Details/INR-015}
  {Automatic summarization}.
\newblock \emph{Foundations and Trends in Information Retrieval}.

\bibitem[{Over and Yen(2004)}]{Over:2004}
Paul Over and James Yen. 2004.
\newblock \href {https://duc.nist.gov/pubs/2004slides/duc2004.intro.pdf} {An
  introduction to {DUC}-2004}.
\newblock \emph{National Institute of Standards and Technology}.

\bibitem[{Paul et~al.(2010)Paul, Zhai, and Girju}]{paul-etal-2010-summarizing}
Michael Paul, ChengXiang Zhai, and Roxana Girju. 2010.
\newblock \href {https://www.aclweb.org/anthology/D10-1007} {Summarizing
  contrastive viewpoints in opinionated text}.
\newblock In \emph{Proceedings of the 2010 Conference on Empirical Methods in
  Natural Language Processing}, pages 66--76, Cambridge, MA. Association for
  Computational Linguistics.

\bibitem[{Rello et~al.(2014)Rello, Saggion, and
  Baeza-Yates}]{rello-etal-2014-keyword}
Luz Rello, Horacio Saggion, and Ricardo Baeza-Yates. 2014.
\newblock \href {https://doi.org/10.3115/v1/W14-1204} {Keyword highlighting
  improves comprehension for people with dyslexia}.
\newblock In \emph{Proceedings of the 3rd Workshop on Predicting and Improving
  Text Readability for Target Reader Populations ({PITR})}, pages 30--37,
  Gothenburg, Sweden. Association for Computational Linguistics.

\bibitem[{Sadeh et~al.(2013)Sadeh, Acquisti, Breaux, Cranor, McDonald,
  Reidenberg, Smith, Liu, Russell, Schaub, and Wilson}]{Sadeh:2013}
Norman Sadeh, Alessandro Acquisti, Travis~D. Breaux, Lorrie~Faith Cranor,
  Aleecia~M. McDonald, Joel~R. Reidenberg, Noah~A. Smith, Fei Liu, N.~Cameron
  Russell, Florian Schaub, and Shomir Wilson. 2013.
\newblock The usable privacy policy project.
\newblock \emph{Technical Report, CMU-ISR-13-119, Carnegie Mellon University}.

\bibitem[{See et~al.(2017)See, Liu, and Manning}]{see-etal-2017-get}
Abigail See, Peter~J. Liu, and Christopher~D. Manning. 2017.
\newblock \href {https://doi.org/10.18653/v1/P17-1099} {Get to the point:
  Summarization with pointer-generator networks}.
\newblock In \emph{Proceedings of the 55th Annual Meeting of the Association
  for Computational Linguistics (Volume 1: Long Papers)}, pages 1073--1083,
  Vancouver, Canada. Association for Computational Linguistics.

\bibitem[{Sharghi et~al.(2018)Sharghi, Borji, Li, Yang, and
  Gong}]{Sharghi:2018}
Aidean Sharghi, Ali Borji, Chengtao Li, Tianbao Yang, and Boqing Gong. 2018.
\newblock \href {https://arxiv.org/abs/1807.10957} {Improving sequential
  determinantal point processes for supervised video summarization}.
\newblock In \emph{Proceedings of the European Conference on Computer Vision
  (ECCV)}.

\bibitem[{Spala et~al.(2018)Spala, Dernoncourt, Chang, and
  Dockhorn}]{spala-etal-2018-web}
Sasha Spala, Franck Dernoncourt, Walter Chang, and Carl Dockhorn. 2018.
\newblock \href {https://www.aclweb.org/anthology/C18-2017} {A web-based
  framework for collecting and assessing highlighted sentences in a document}.
\newblock In \emph{Proceedings of the 27th International Conference on
  Computational Linguistics: System Demonstrations}, pages 78--81, Santa Fe,
  New Mexico. Association for Computational Linguistics.

\bibitem[{Stanovsky et~al.(2016)Stanovsky, Ficler, Dagan, and
  Goldberg}]{stanovsky2016getting}
Gabriel Stanovsky, Jessica Ficler, Ido Dagan, and Yoav Goldberg. 2016.
\newblock \href {http://arxiv.org/abs/1603.01648} {Getting more out of syntax
  with props}.

\bibitem[{Tan et~al.(2017)Tan, Wan, and Xiao}]{tan-etal-2017-abstractive}
Jiwei Tan, Xiaojun Wan, and Jianguo Xiao. 2017.
\newblock \href {https://doi.org/10.18653/v1/P17-1108} {Abstractive document
  summarization with a graph-based attentional neural model}.
\newblock In \emph{Proceedings of the 55th Annual Meeting of the Association
  for Computational Linguistics (Volume 1: Long Papers)}, pages 1171--1181,
  Vancouver, Canada. Association for Computational Linguistics.

\bibitem[{Vanderwende et~al.(2007)Vanderwende, Suzuki, Brockett, and
  Nenkova}]{Vanderwende:2007}
Lucy Vanderwende, Hisami Suzuki, Chris Brockett, and Ani Nenkova. 2007.
\newblock \href {https://www.cis.upenn.edu/~nenkova/papers/ipm.pdf} {Beyond
  {SumBasic}: {T}ask-focused summarization with sentence simplification and
  lexical expansion}.
\newblock \emph{Information Processing and Management}, 43(6):1606--1618.

\bibitem[{Vladutz(1983)}]{vladutz-1983-natural}
G.~Vladutz. 1983.
\newblock \href {https://doi.org/10.3115/974194.974221} {Natural language text
  segmentation techniques applied to the automatic compilation of printed
  subject indexes and for online database access}.
\newblock In \emph{First Conference on Applied Natural Language Processing},
  pages 136--142, Santa Monica, California, USA. Association for Computational
  Linguistics.

\bibitem[{Woodsend and Lapata(2010)}]{woodsend-lapata-2010-automatic}
Kristian Woodsend and Mirella Lapata. 2010.
\newblock \href {https://www.aclweb.org/anthology/P10-1058} {Automatic
  generation of story highlights}.
\newblock In \emph{Proceedings of the 48th Annual Meeting of the Association
  for Computational Linguistics}, pages 565--574, Uppsala, Sweden. Association
  for Computational Linguistics.

\bibitem[{Yang et~al.(2019)Yang, Dai, Yang, Carbonell, Salakhutdinov, and
  Le}]{NIPS2019_8812}
Zhilin Yang, Zihang Dai, Yiming Yang, Jaime Carbonell, Russ~R Salakhutdinov,
  and Quoc~V Le. 2019.
\newblock \href
  {http://papers.nips.cc/paper/8812-xlnet-generalized-autoregressive-pretraining-for-language-understanding.pdf}
  {Xlnet: Generalized autoregressive pretraining for language understanding}.
\newblock In \emph{Advances in Neural Information Processing Systems 32}, pages
  5753--5763. Curran Associates, Inc.

\end{thebibliography}
\bibliographystyle{acl_natbib}

\appendix

\section{Example System Outputs}

We present example system outputs contrasting our highlighting method with traditional sentence extraction and human abstraction. 
Highlighting helps readers quickly skim through a large amount of text to grasp the main points. 
We observe that the XLNet segments are better than those obtained using the subtree method—not only can they aid reader comprehension but they are also self-contained and more concise. 
Further, we show example text segments produced by our XLNet algorithm, accompanied by their scores of self-containedness judged by five human evaluators, whose average scores are reported. 
Results of human evaluation suggest that text segments produced by our model demonstrate a high degree of self-containedness.

\clearpage

\begin{table}[!h]
\setlength{\tabcolsep}{5pt}
\renewcommand{\arraystretch}{1.1}
\begin{scriptsize}
\begin{minipage}{\textwidth}
\centering
\textsf{
\begin{tabular}[t]{|p{6.1in}|}
\hline
\textbf{Human Abstract}\Tstrut\\[1mm]
$\bullet$ Exxon and Mobil discuss combining business operations.\\[1mm]
$\bullet$ A possible Exxon-Mobil merger would reunite 2 parts of Standard Oil broken up by the Supreme Court in 1911.\\[1mm]
$\bullet$ Low crude oil prices and the high cost of exploration are motives for a merger that would create the world's largest oil company.\\[1mm]
$\bullet$ As Exxon-Mobil merger talks continue, stocks of both companies surge.\\[1mm]
$\bullet$ The merger talks show that corporate mergers are back in vogue.\\[1mm]
$\bullet$ Antitrust lawyers, industry analysts, and government officials say a merger would require divestitures.\\[1mm]
$\bullet$ A Mobil employee worries that a merger would put thousands out of work, but notes that his company's stock would go up.\\[1mm]
\hline\hline
\textbf{Extractive Summary}\Tstrut\\[1mm]
$\bullet$ The boards of Exxon Corp. and Mobil Corp. are expected to meet Tuesday to consider a possible merger agreement that would form the world's largest oil company, a source close to the negotiations said Friday.\\[1mm] 
$\bullet$ Exxon and Mobil, the nation's two largest oil companies, confirmed Friday that they were discussing a possible merger, and antitrust lawyers, industry analysts and government officials predicted that any deal would require the sale of important large pieces of such a new corporate behemoth.\\[1mm] 
$\bullet$ The reported talks between Exxon, whose annual revenue exceeds that of General Electric Co., and Mobil, the No. 2 U.S. oil company, came as oil prices sank to their lowest in almost 12 years.\\[1mm] 
\hline\hline
\textbf{Highlighting (Tree Segments)}\Tstrut\\[1mm]
$\bullet$ \textcolor{dodgerblue}{Whether or not the talks between Exxon and Mobil lead to a merger or some other business combination}, America's economic history is already being rewritten.\\[1mm]
$\bullet$ \textcolor{dodgerblue}{The boards of Exxon Corp. and Mobil Corp. are expected to meet Tuesday to consider a possible merger agreement that would form the world's largest oil company}, a source close to the negotiations said Friday. \\[1mm]
$\bullet$ Exxon Corp. and Mobil Corp. \textcolor{dodgerblue}{have held discussions about combining their business operations, a person involved in the talks said Wednesday}.\\[1mm]
$\bullet$ News \textcolor{dodgerblue}{that Exxon and Mobil, two giants in the energy patch, were in merger talks last week} is the biggest sign yet that corporate marriages are back in vogue. \\[1mm] 
$\bullet$ \textcolor{dodgerblue}{Shares of Exxon, the biggest U.S. oil company,} rose \$1.6875, or 2.3 percent, to \$74.375.\\[1mm]
$\bullet$ Some analysts \textcolor{dodgerblue}{said that if the two giants reached an agreement, it was likely to be in the form of a takeover by Exxon of Mobil}.\\[1mm]
$\bullet$ Exxon was then known as Standard Oil of New Jersey, and \textcolor{dodgerblue}{Mobil consisted of two companies: Standard Oil of New York and Vacuum Oil}.\\[1mm]
$\bullet$ Which is \textcolor{dodgerblue}{why Mobil and Exxon are considering combining into the world's largest oil company}.\\[1mm]
\hline\hline
\textbf{Highlighting (XLNet Segments)}\Tstrut\\[1mm]
$\bullet$ \textcolor{fuchsia}{Whether or not the talks between Exxon and Mobil lead to a merger} or some other business combination, America's economic history is already being rewritten.\\[1mm]
$\bullet$ Still, it boggles the mind to accept \textcolor{fuchsia}{the notion that hardship is driving profitable Big Oil to either merge}, as British Petroleum and Amoco have already agreed to do, or at least to consider the prospect, as Exxon and Mobil are doing.\\[1mm]
$\bullet$ Oil stocks led the way as investors soaked up the news of continuing talks between \textcolor{fuchsia}{Exxon and Mobil on a merger that would create the world's largest oil company}.\\[1mm]
$\bullet$ Although the companies only confirmed that they were discussing the possibility of a merger, \textcolor{fuchsia}{a person close to the discussions said the boards of both Exxon and Mobil} were expected to meet Tuesday to consider an agreement.\\[1mm]
$\bullet$ Analysts predicted \textcolor{fuchsia}{that there would be huge cuts in duplicate staff from both companies}, which employ 122,700 people.\\[1mm]
$\bullet$ They said \textcolor{fuchsia}{the transaction would probably be an exchange of Mobil shares for Exxon shares}.\\[1mm]
$\bullet$ But \textcolor{fuchsia}{this has been a particularly unsettling year for the oil industry}, and there is little prospect that crude oil prices will recover soon.\\[1mm]
$\bullet$ The merger discussions come against a backdrop of particularly severe pressure on Lucio Noto, the chairman, president and chief executive of Mobil, to find new reserves of oil and natural gas and to keep big projects \textcolor{fuchsia}{profitable at a time of a deep decline in crude oil prices}.\\[1mm]
$\bullet$ If there is a reason this merger might get extra attention, it will be \textcolor{fuchsia}{because Exxon and Mobil have not been terribly friendly toward either the Clinton administration's or the European Union's positions on global warming}.\\[1mm]
\hline
\end{tabular}}
\caption{
Example system outputs for a topic in DUC-04. 
Highlighting allows readers to quickly sift through a large amount of text to grasp the main points. 
XLNet segments perform better than tree segments. Not only can they aid reader comprehension but they are also self-contained and more concise. 
}
\end{minipage}
\hfill
\end{scriptsize}
\label{tab:results-output}
\vspace{-0.1in}
\end{table}

\begin{table*}[h!]
\setlength{\tabcolsep}{5pt}
\renewcommand{\arraystretch}{1.1}
\begin{scriptsize}
\begin{minipage}[b]{\hsize} 
\centering
\textsf{
\begin{tabular}[t]{|p{6.1in}|} 
\hline
\textbf{Human Abstract}\Tstrut\\[1mm]
$\bullet$ After years of civil war, Congo in October 1998 was again in turmoil as rebel forces fought to overthrow the government of President Kabila.\\[1mm]
$\bullet$ The rebels, ethnic Tutsis, disenchanted members of Kabila's army and his political opponents, were said to be supported by Rwandan and Ugandan forces while Kabila was backed by Angola, Zimbabwe, Namibia, Sudan and Ugandan rebels.\\[1mm]
$\bullet$ Ugandan forces while Kabila was backed by Angola, Zimbabwe, Namibia, Sudan and Ugandan rebels.\\[1mm]
$\bullet$ At first the rebels advanced to the outskirts of the capital, Kinshasa, but foreign troops pushed them back to the extreme eastern part of the country.\\[1mm]
$\bullet$ The rebels then launched a counter offensive but by mid-October it was not clear who would prevail.\\[1mm]
\hline\hline
\textbf{Extractive Summary}\Tstrut\\[1mm]
$\bullet$ After a day of fighting, Congolese rebels said Sunday they had entered Kindu, the strategic town and airbase in eastern Congo used by the government to halt their advances.\\[1mm] 
$\bullet$ Rebels in eastern Congo on Saturday said they shot down a passenger jet ferrying 40 government soldiers into a strategic airport facing a rebel assault.\\[1mm] 
$\bullet$ A rebel defeat, on the other hand, would put the coalition of ethnic Tutsis, disenchanted members of the Congolese army and opposition politicians on the defensive and give a boost to Kabila's efforts to fend off the rebellion launched Aug. 2. Rebel commander Richard Mondo said troops had fired artillery rounds into Kindu Monday and early Tuesday, sending the population fleeing out of town.\\[1mm] 
$\bullet$ On Saturday, the rebels said they shot down a Congolese Boeing 727 which was attempting to land at Kindu air base with 40 troops and ammunition.\\[1mm]
\hline\hline
\textbf{Highlighting (Tree Segments)}\Tstrut\\[1mm]
$\bullet$ \textcolor{dodgerblue}{Rebels attacked a village in western Uganda and killed six civilians before soldiers drove them off}, a military spokesman said Thursday.\\[1mm]
$\bullet$ Congolese rebels \textcolor{dodgerblue}{have taken their two-month campaign to oust President Laurent Kabila to the Internet}.\\[1mm]
$\bullet$ \textcolor{dodgerblue}{A day after shooting down a jetliner, Congolese rebels and their Rwandan allies} pushed Sunday through government defense lines, showing the confidence of a victor in a week-old battle for a strategic air base.\\[1mm]
$\bullet$ After a day of fighting, Congolese rebels \textcolor{dodgerblue}{said Sunday they had entered Kindu, the strategic town and airbase in eastern Congo used by the government to halt their advances}.\\[1mm]
$\bullet$ Rebels in eastern Congo on Saturday \textcolor{dodgerblue}{said they shot down a passenger jet ferrying 40 government soldiers into a strategic airport facing a rebel assault}.\\[1mm]
$\bullet$ A day after shooting down a jetliner carrying 40 people, rebels \textcolor{dodgerblue}{clashed with government troops near a strategic airstrip in eastern Congo on Sunday}.\\[1mm]
$\bullet$ Kabila has turned \textcolor{dodgerblue}{Kindu into a launching pad for a counteroffensive against rebel positions in eastern Congo}.\\[1mm]
\hline\hline
\textbf{Highlighting (XLNet Segments)}\Tstrut\\[1mm]
$\bullet$ Congolese rebels \textcolor{fuchsia}{have taken their two-month campaign to oust President Laurent Kabila} to the Internet.\\[1mm]
$\bullet$ \textcolor{fuchsia}{The bloody bandages of injured rebels trucked back to this rear base} Wednesday offered evidence that the three-day battle for the strategic air base at Kindu was not going well for those fighting to oust Congolese President Laurent Kabila.\\[1mm]
$\bullet$ Rebels in eastern Congo on Saturday said \textcolor{fuchsia}{they shot down a passenger jet ferrying 40 government soldiers} into a strategic airport facing a rebel assault.\\[1mm]
$\bullet$ After trekking several hundred kilometers through dense tropical forest, \textcolor{fuchsia}{thousands of rebel fighters have gathered 19 kilometers outside Kindu}, where troops loyal to President Laurent Kabila have used an air base as a launching pad for offensives.\\[1mm]
$\bullet$ On Saturday, the rebels \textcolor{fuchsia}{said they shot down a Congolese Boeing 727 which was attempting to land at Kindu air base} with 40 troops and ammunition.\\[1mm]
$\bullet$ President Yoweri Museveni insists they will remain there until Ugandan security is guaranteed, despite Congolese President \textcolor{fuchsia}{Laurent Kabila's protests that Uganda is backing Congolese rebels} attempting to topple him.\\[1mm]
$\bullet$ \textcolor{fuchsia}{The rebels see Kindu as a major prize in their two-month revolt} against President Laurent Kabila, whom they accuse of mismanagement, corruption and warmongering among Congo's 400 tribes.\\[1mm]
$\bullet$ \textcolor{fuchsia}{Both countries say they have legitimate security interests in eastern Congo and accuse Kabila of} \textcolor{mediumseagreen}{failing to rid the common border area} of Rwandan and Ugandan rebels.\\[1mm]
$\bullet$ \textcolor{fuchsia}{The rebels say they now control one-third of Kindu and are poised to overrun the rest} of the town.\\[1mm]
\hline
\end{tabular}}
\end{minipage}
\hfill
\end{scriptsize}
\caption{
Example system outputs for a topic in DUC-04. 
Highlighting allows readers to quickly sift through a large amount of text to grasp the main points. 
XLNet segments perform better than tree segments. Not only can they aid reader comprehension but they are also self-contained and more concise. 
Our method further allows multiple segments, denoted by \legendsquare{fuchsia} and \legendsquare{mediumseagreen}, to be selected from the same sentence.
}
\label{tab:results-output}
\vspace{-0.1in}
\end{table*}

\begin{table*}[h!]
\setlength{\tabcolsep}{5pt}
\renewcommand{\arraystretch}{1.1}
\begin{scriptsize}
\begin{minipage}[b]{\hsize}
\centering
\textsf{
\begin{tabular}[t]{|p{6.1in}|}
\hline
\textbf{Human Abstract}\Tstrut\\[1mm]
$\bullet$ Eleven countries were to adopt a common European currency, the euro, on Dec. 31, 1998.\\[1mm]
$\bullet$ In November and December there were various reactions.\\[1mm]
$\bullet$ France made moves toward a pan-European equity market.\\[1mm]
$\bullet$ Ten of the countries quickly cut interest rates causing fear of overheating in some economies.\\[1mm]
$\bullet$ In Denmark, which had earlier rejected the euro, a majority was now in favor.\\[1mm]
$\bullet$ And in faraway China, the euro was permitted in financial exchanges.\\[1mm]
$\bullet$ Whatever the outcome, the euro's birthday, Dec. 31, 1998, would be an historical date.\\[1mm]
$\bullet$ Some saw it as a step towards political union while others already considered themselves as citizens of Europe.\\[1mm]
\hline\hline
\textbf{Extractive Summary}\Tstrut\\[1mm]
$\bullet$ In a surprise move, nations adopting the new European currency, the euro, dropped key interest rates Thursday, effectively setting the rate that will be adopted throughout the euro zone on Jan. 1.\\[1mm] 
$\bullet$ The annual inflation rate in the 11 nations that adopt the euro as their shared currency on Jan. 1 fell to 0.9 percent in November, the European Union's statistics agency reported Wednesday.\\[1mm] 
$\bullet$ Wim Duisenberg, the head of the new European Central Bank, said in an interview published Wednesday that he won't step down after completing half his term as earlier agreed.\\[1mm] 
$\bullet$ Ten of the 11 countries adopting the euro dropped their interest rate to 3 percent.\\[1mm]
$\bullet$ Duisenberg was named this spring as head of the new European Central Bank, which will govern the policies of the euro, the new single currency which goes into effect Jan. 1.\\[1mm]
\hline\hline
\textbf{Highlighting (Tree Segments)}\Tstrut\\[1mm]
$\bullet$ \textcolor{dodgerblue}{Two days before the new euro currency goes into effect for 11 European Union members}, a growing number of Danes believe their country should take part, according to a poll published Tuesday.\\[1mm]
$\bullet$ \textcolor{dodgerblue}{Wim Duisenberg, the head of the new European Central Bank,} said in an interview published Wednesday that he won't step down after completing half his term as earlier agreed.\\[1mm]
$\bullet$ In a surprise move, \textcolor{dodgerblue}{nations adopting the new European currency, the euro,} dropped key interest rates Thursday, effectively setting the rate that will be adopted throughout the euro zone on Jan. 1.\\[1mm]
$\bullet$ Making their first collective decision about monetary policy, \textcolor{dodgerblue}{the 11 European nations launching a common currency on Jan. 1 cut interest rates Thursday in a surprise move that won market confidence}.\\[1mm]
$\bullet$ In a surprise move, nations adopting the new European currency, the euro, \textcolor{dodgerblue}{dropped key interest rates Thursday, effectively setting the rate that will be adopted throughout the euro zone on Jan. 1}.\\[1mm]
$\bullet$ China made trading in the euro official Monday, \textcolor{dodgerblue}{announcing authorization for the European common currency's use in trade and financial dealings starting Jan. 1}.\\[1mm]
$\bullet$ \textcolor{dodgerblue}{The annual inflation rate in the 11 nations that adopt the euro as their shared currency on Jan. 1} fell to 0.9 percent in November, the European Union's statistics agency reported Wednesday.\\[1mm]
$\bullet$ The year 1999 is the official start-up date \textcolor{dodgerblue}{of the euro, the common European currency that will unite 11 countries monetarily}.\\[1mm]
\hline\hline
\textbf{Highlighting (XLNet Segments)}\Tstrut\\[1mm]
$\bullet$ \textcolor{fuchsia}{Two days before the new euro currency goes into effect} for 11 European Union members, a growing number of Danes believe their country should take part, according to a poll published Tuesday.\\[1mm]
$\bullet$ Making their first collective decision about monetary policy, the \textcolor{fuchsia}{11 European nations launching a common currency on Jan. 1 cut} interest rates Thursday in a surprise move that won market confidence.\\[1mm]
$\bullet$ \textcolor{fuchsia}{The annual inflation rate in the 11 nations that adopt the euro as their shared currency on Jan. 1 fell to 0.9 percent} in November, the European Union's statistics agency reported Wednesday.\\[1mm]
$\bullet$ French \textcolor{fuchsia}{authorities said Thursday that the Paris stock exchange would join an alliance between London and Frankfurt that is seen as the precursor of a pan-European market}.\\[1mm]
$\bullet$ Ten of the 11 countries joining \textcolor{fuchsia}{European economic union dropped their key interest rate to 3 percent}, with Portugal making the most significant plunge, from 3.75 percent.\\[1mm]
$\bullet$ Not only that, the notion of \textcolor{fuchsia}{a Europe-wide exchange raises technical questions about the compatibility not just of trading systems but also of the regulations governing stock market operations in the countries that will use the euro} beginning in January.\\[1mm]
\hline
\end{tabular}}
\end{minipage}
\hfill
\end{scriptsize}
\caption{
Example system outputs for a topic in DUC-04. 
Highlighting allows readers to quickly sift through a large amount of text to grasp the main points. 
XLNet segments perform better than tree segments. Not only can they aid reader comprehension but they are also self-contained and more concise. 
}
\label{tab:results-output}
\vspace{-0.1in}
\end{table*}

\begin{table*}[h!]
\setlength{\tabcolsep}{5pt}
\renewcommand{\arraystretch}{1.1}
\begin{scriptsize}
\begin{minipage}[b]{\hsize}
\centering
\textsf{
\begin{tabular}[t]{|p{6.1in}|}
\hline
\textbf{Human Abstract}\Tstrut\\[1mm]
$\bullet$ Boeing 737-400 plane with 102 people on board crashed into a mountain in the West Sulawesi province of Indonesia, on Monday, January 01, 2007, killing at least 90 passengers, with 12 possible survivors.\\[1mm]
$\bullet$ The plane was Adam Air flight KI-574, departing at 12:59 pm from Surabaya on Java bound for Manado in northeast Sulawesi.\\[1mm]
$\bullet$ There were three Americans on board, it is not know if they survived.\\[1mm]
$\bullet$ The cause of the crash is not known at this time but it is possible bad weather was a factor.\\[1mm]
\hline\hline
\textbf{Extractive Summary}\Tstrut\\[1mm]
$\bullet$ Three Americans were among the 102 passengers and crew on board an Adam Air plane which crashed into a remote mountainous region of Indonesia, an airline official said Tuesday.\\[1mm] 
$\bullet$ Rescue teams Tuesday found the smoldering wreckage of an Indonesian jetliner that went missing over Indonesia's Sulawesi island during a storm.\\[1mm] 
$\bullet$ The Indonesian rescue team Tuesday arrived at the mountainous area in West Sulawesi province where a passenger plane with 102 people onboard crashed Monday, finding at least 90 bodies at the scene.\\[1mm] 
$\bullet$ The Indonesian Navy (TNI AL) has sent two Cassa planes to carry the bodies of five of its members who were killed in a plane crash in the Indonesian island of Sulawesi late Monday.\\[1mm]
\hline\hline
\textbf{Highlighting (Tree Segments)}\Tstrut\\[1mm]
$\bullet$ \textcolor{dodgerblue}{An Indonesian passenger plane carrying 102 people disappeared in stormy weather on Monday}, and rescue teams were sent to search an area where military aviation officials feared the Boeing 737-400 aircraft may have crashed.\\[1mm]
$\bullet$ \textcolor{dodgerblue}{Indonesian Transportation Ministry' s air transportation director general M. Ichsan Tatang said the weather in Polewali of Sulaweisi province} was bad when the plane took off from Surabaya.\\[1mm]
$\bullet$ Three Americans \textcolor{dodgerblue}{were among the 102 passengers and crew on board an Adam Air plane which crashed into a remote mountainous region of Indonesia}, an airline official said Tuesday.\\[1mm]
$\bullet$ An Indonesian passenger plane carrying 102 people disappeared in stormy weather on Monday, and \textcolor{dodgerblue}{rescue teams were sent to search an area where military aviation officials feared the Boeing 737-400 aircraft may have crashed}.\\[1mm]
$\bullet$ Rescue teams Tuesday found \textcolor{dodgerblue}{the smoldering wreckage of an Indonesian jetliner that went missing over Indonesia's Sulawesi island during a storm}, officials said.\\[1mm]
$\bullet$ Chinese Foreign Minister Li Zhaoxing on Tuesday \textcolor{dodgerblue}{sent a message of condolences to his Indonesian counterpart Hassan Wirayuda over Monday's plane crash}.\\[1mm]
\hline\hline
\textbf{Highlighting (XLNet Segments)}\Tstrut\\[1mm]
$\bullet$ \textcolor{fuchsia}{An Indonesian passenger plane carrying 102 people disappeared in stormy weather on Monday}, and rescue teams were sent to search an area where military aviation officials feared the Boeing 737-400 aircraft may have crashed.\\[1mm]
$\bullet$ Three Americans were among the 102 passengers and crew on board an \textcolor{fuchsia}{Adam Air plane which crashed into a remote mountainous region of} Indonesia, an airline official said Tuesday.\\[1mm]
$\bullet$ Indonesian \textcolor{fuchsia}{President Susilo Bambang Yudhoyono said Tuesday he was deeply concerned with the crash of} a passenger plane and the sinking of a ferry in the last few days that might have killed hundreds of people.\\[1mm]
$\bullet$ \textcolor{fuchsia}{Chinese Foreign Minister Li Zhaoxing on Tuesday sent a message} of condolences to his Indonesian counterpart Hassan Wirayuda over Monday's plane crash.\\[1mm]
$\bullet$ The Indonesian Navy (TNI AL) has sent two Cassa planes to carry the bodies of five of its members \textcolor{fuchsia}{who were killed in a plane crash in the Indonesian} island of Sulawesi late Monday.\\[1mm]
$\bullet$ An Indonesian passenger plane carrying 102 people disappeared in stormy weather on Monday, and rescue teams were sent to search an area where \textcolor{fuchsia}{military aviation officials feared the Boeing 737-400 aircraft may have crashed}.\\[1mm]
$\bullet$ In the message, \textcolor{fuchsia}{Li said he was "shocked" to learn} of the tragedy and expressed deep condolences to the victims of the accident.\\[1mm]
$\bullet$ In 1960s, some planes and helicopters crashed on Masalombo area after they were absorbed by air pockets.
Martono likened \textcolor{fuchsia}{Masalombo area to Bermuda Triangle where many ships and airplanes went missing}.\\[1mm]
$\bullet$ \textcolor{fuchsia}{Latest reports said at least 12 passengers including five children survived} the accident but they were in critical condition and sent to a nearby hospital in Polewali.\\[1mm]
\hline
\end{tabular}}
\end{minipage}
\hfill
\end{scriptsize}
\caption{
Example system outputs for a topic in TAC-11. 
Highlighting allows readers to quickly sift through a large amount of text to grasp the main points. 
XLNet segments perform better than tree segments. Not only can they aid reader comprehension but they are also self-contained and more concise. 
}
\label{tab:results-output}
\vspace{-0.1in}
\end{table*}

\begin{table*}[h!]
\setlength{\tabcolsep}{5pt}
\renewcommand{\arraystretch}{1.1}
\begin{scriptsize}
\begin{minipage}[b]{\hsize} 
\centering
\textsf{
\begin{tabular}[t]{|p{6.1in}|}
\hline
\textbf{Human Abstract}\Tstrut\\[1mm]
$\bullet$ Internet security needs a global approach because it is a global problem.\\[1mm]
$\bullet$ Pakistan tried to block a riot-sparking video and accidentally blocked world YouTube access.\\[1mm]
$\bullet$ Internet sabotage shut down digital infrastructure in Estonia and Bangladesh.\\[1mm]
$\bullet$ Overseas hackers accessed confidential information from South Korea.\\[1mm]
$\bullet$ China and Taiwan are both accused of Internet attacks to steal secret data.\\[1mm]
$\bullet$ The U.S. considered including cyberspace regulation in rules of international warfare.\\[1mm]
$\bullet$ South Korea's real-name system authenticates identity information on applications for online accounts.\\[1mm]
$\bullet$ The UAE is establishing a computer emergency response team.\\[1mm]
$\bullet$ Computer whizzes sell security vulnerability information to both software vendors and criminals.\\[1mm]
\hline\hline
\textbf{Extractive Summary}\Tstrut\\[1mm]
$\bullet$ Telecoms and computer executives, legal officials and UN agencies on Friday warned that the world needs to take a global approach to tackling cybercrime and security issues on the Internet.\\[1mm] 
$\bullet$ Taiwan's Internet market has matured over the past 10 years, but the ratio of Internet users worried about Internet security has risen significantly, according the results of a telephone survey released Sunday by the Ministry of Transportation and Communications (MOTC).\\[1mm] 
$\bullet$ The National Security Bureau (NSB) has never permitted hacking activities nor any other attack on computer and Internet systems at home or abroad, the NSB said in a news release issued Thursday.\\[1mm] 
$\bullet$ Since the adoption by the South Korean government in 2005 of the Internet real- name system, people's privacy, reputation and economic rights are better protected, according to the Ministry of Information and Telecommunication.\\[1mm]
\hline\hline
\textbf{Highlighting (Tree Segments)}\Tstrut\\[1mm]
$\bullet$ \textcolor{dodgerblue}{South Korea's presidential mansion, the Blue House, has come under cyber-attack from overseas hackers who accessed some confidential information}, officials said Tuesday.\\[1mm]
$\bullet$ \textcolor{dodgerblue}{The National Security Bureau (NSB) has never permitted hacking activities nor any other attack on computer and Internet systems at home} or abroad, the NSB said in a news release issued Thursday.\\[1mm]
$\bullet$ \textcolor{dodgerblue}{Since the adoption by the South Korean government in 2005 of the Internet real- name system}, people's privacy, reputation and economic rights are better protected, according to the Ministry of Information and Telecommunication.\\[1mm]
$\bullet$ Telecoms and computer executives, legal officials and UN agencies on Friday \textcolor{dodgerblue}{warned that the world needs to take a global approach to tackling cybercrime and security issues on the Internet}.\\[1mm]
$\bullet$ Bangladesh on Tuesday launched an investigation \textcolor{dodgerblue}{after the country's Internet link was sabotaged, disrupting communications nationwide for most of the day}.\\[1mm]
$\bullet$ Since the adoption by the South Korean government in 2005 of the Internet real- name system, people's privacy, reputation and economic rights \textcolor{dodgerblue}{are better protected, according to the Ministry of Information and Telecommunication}.\\[1mm]
\hline\hline
\textbf{Highlighting (XLNet Segments)}\Tstrut\\[1mm]
$\bullet$ \textcolor{fuchsia}{Taiwan's Internet market has matured over the past 10 years}, but the ratio of Internet users worried about Internet security has risen significantly, according the results of a telephone survey released Sunday by the Ministry of Transportation and Communications (MOTC).\\[1mm]
$\bullet$ Since the adoption by the \textcolor{fuchsia}{South Korean government in 2005 of the Internet real- name system}, people's privacy, reputation and economic rights are better protected, according to the Ministry of Information and Telecommunication.\\[1mm]
$\bullet$ Telecoms and computer executives, legal officials and \textcolor{fuchsia}{UN agencies on Friday warned that the world needs to take} a global approach to tackling cybercrime and security issues on the Internet.\\[1mm]
$\bullet$ \textcolor{fuchsia}{The attacks were discovered about two weeks after they happened when the entire computer network underwent a security check in} early March, the Blue House said in a statement.\\[1mm]
$\bullet$ \textcolor{fuchsia}{The National Security Bureau (NSB) has never permitted hacking activities} nor any other attack on computer and Internet systems at home or abroad, the NSB said in a news release issued Thursday.\\[1mm]
$\bullet$ \textcolor{fuchsia}{When Estonian authorities began removing a bronze statue of a World War II-era Soviet soldier} from a park in this bustling Baltic seaport last month, they expected violent street protests by Estonians of Russian descent.\\[1mm]
$\bullet$ Cox News Service WASHINGTON -- The United States must protect \textcolor{fuchsia}{its interests in cyberspace and outer space against threats from} China and other nations, Sen. Bill Nelson said at a hearing Wednesday.\\[1mm]
$\bullet$ Taiwan's Internet market has matured over the past 10 years, but \textcolor{fuchsia}{the ratio of Internet users worried about Internet security has risen significantly}, according the results of a telephone survey released Sunday by the Ministry of Transportation and Communications (MOTC).\\[1mm]
$\bullet$ What \textcolor{fuchsia}{followed was what some here describe as the first war in cyberspace}, a monthlong campaign that has forced Estonian authorities to defend their pint-size Baltic nation from a data flood that they say was set off by orders from Russia or ethnic Russian sources in retaliation for the removal of the statue.\\[1mm]
\hline
\end{tabular}}
\end{minipage}
\hfill
\end{scriptsize}
\caption{
Example system outputs for a topic in TAC-11. 
Highlighting allows readers to quickly sift through a large amount of text to grasp the main points. 
XLNet segments perform better than tree segments. Not only can they aid reader comprehension but they are also self-contained and more concise. 
}
\label{tab:results-output}
\vspace{-0.1in}
\end{table*}

\begin{table*}[h!]
\setlength{\tabcolsep}{0pt}
\renewcommand{\arraystretch}{1.1}
\begin{scriptsize}
\begin{minipage}[b]{\hsize} 
\centering
\textsf{
\begin{tabular}[t]{p{6.2in}} 
\textbf{\textcolor{blue}{[Original Sentence]} District Attorney David Roger agreed to drop charges including kidnapping, armed robbery, assault with a deadly weapon and conspiracy against both men.}\\[1mm]
\toprule
$\bullet$ District \textcolor{orangered}{Attorney David Roger agreed to drop charges including kidnapping}, armed robbery, assault with a deadly weapon and conspiracy against both men. (4.0)\\[1mm]
$\bullet$ District \textcolor{darkgreen}{Attorney David Roger agreed to drop charges} including kidnapping, armed robbery, assault with a deadly weapon and conspiracy against both men. (3.8)\\[1mm]
$\bullet$ District Attorney David Roger agreed to drop charges including kidnapping, armed robbery, \textcolor{dodgerblue}{assault with a deadly weapon and conspiracy} against both men. (3.6)\\[1mm]
\bottomrule
\end{tabular}}
\end{minipage}
\hfill
\end{scriptsize}
\caption{
Example text segments produced by the XLNet model. 
The scores of \emph{self-containedness} are shown in parentheses. 
Each segment is judged by five human evaluators on a scale of 1 (worst) to 5 (best) and we report their average scores. 
Human evaluation suggests that text segments generated by our model demonstrate a high degree of self-containedness.
} 
\label{tab:results-output}
\end{table*}

\begin{table*}[h!]
\setlength{\tabcolsep}{0pt}
\renewcommand{\arraystretch}{1.1}
\begin{scriptsize}
\begin{minipage}[b]{\hsize} 
\centering
\textsf{
\begin{tabular}[t]{p{6.2in}} 
\textbf{\textcolor{blue}{[Original Sentence]} “I can't imagine anyone saying no,” the 21-year-old college student said last week as, teary-eyed, she met 8-month-old Allison Brown, carefully cuddling the wide-eyed baby so as not to bump each other's healing incisions.}\\[1mm]
\toprule
$\bullet$ “I can't \textcolor{orangered}{imagine anyone saying no,” the 21-year-old college student said} last week as, teary-eyed, she met 8-month-old Allison Brown, carefully cuddling the wide-eyed baby so as not to bump each other's healing incisions. (3.0)\\[1mm]
$\bullet$ “I can't imagine anyone saying no,” the 21-year-old college student said last week as, teary-eyed, \textcolor{darkgreen}{she met 8-month-old Allison Brown}, carefully cuddling the wide-eyed baby so as not to bump each other's healing incisions. (3.8)\\[1mm]
$\bullet$ “I can't imagine anyone saying no,” the 21-year-old college student said last week as, teary-eyed, she met 8-month-old Allison Brown, carefully cuddling \textcolor{dodgerblue}{the wide-eyed baby so as not to bump each other's healing incisions}. (3.6)\\[1mm]
$\bullet$ “I can't imagine anyone saying no,” the 21-year-old college student said last week as, teary-eyed, she met 8-month-old Allison Brown, \textcolor{fuchsia}{carefully cuddling the wide-eyed baby so as not to bump} each other's healing incisions. (4.0)\\[1mm]
\bottomrule
\end{tabular}}
\end{minipage}
\hfill
\end{scriptsize}
\caption{
Example text segments produced by the XLNet model. 
The scores of \emph{self-containedness} are shown in parentheses. 
Each segment is judged by five human evaluators on a scale of 1 (worst) to 5 (best) and we report their average scores. 
Human evaluation suggests that text segments generated by our model demonstrate a high degree of self-containedness.
} 
\label{tab:results-output}
\end{table*}

\begin{table*}[h!]
\setlength{\tabcolsep}{0pt}
\renewcommand{\arraystretch}{1.1}
\begin{scriptsize}
\begin{minipage}[b]{\hsize} 
\centering
\textsf{
\begin{tabular}[t]{p{6.2in}} 
\textbf{\textcolor{blue}{[Original Sentence]} Madoff is charged with stealing as much as \$50 billion, in part to cover a pattern of massive losses, even as he cultivated a reputation as a financial mastermind and prominent philanthropist.}\\[1mm]
\toprule
$\bullet$ \textcolor{orangered}{Madoff is charged with stealing} as much as \$50 billion, in part to cover a pattern of massive losses, even as he cultivated a reputation as a financial mastermind and prominent philanthropist. (3.6)\\[1mm]
$\bullet$ Madoff is charged with stealing as much as \$50 billion, in \textcolor{darkgreen}{part to cover a pattern of} massive losses, even as he cultivated a reputation as a financial mastermind and prominent philanthropist. (2.0)\\[1mm]
$\bullet$ Madoff is charged with stealing as much as \$50 billion, in part to cover a pattern of massive losses, \textcolor{dodgerblue}{even as he cultivated a reputation as} a financial mastermind and prominent philanthropist. (3.0)\\[1mm]
$\bullet$ Madoff is charged with stealing as much as \$50 billion, in part to cover a pattern of massive losses, even as \textcolor{fuchsia}{he cultivated a reputation as a financial mastermind and prominent philanthropist}. (2.8)\\[1mm]
\bottomrule
\end{tabular}}
\end{minipage}
\hfill
\end{scriptsize}
\caption{
Example text segments produced by the XLNet model. 
The scores of \emph{self-containedness} are shown in parentheses. 
Each segment is judged by five human evaluators on a scale of 1 (worst) to 5 (best) and we report their average scores. 
Human evaluation suggests that text segments generated by our model demonstrate a high degree of self-containedness.
} 
\label{tab:results-output}
\end{table*}

\begin{table*}[h!]
\setlength{\tabcolsep}{0pt}
\renewcommand{\arraystretch}{1.1}
\begin{scriptsize}
\begin{minipage}[b]{\hsize} 
\centering
\textsf{
\begin{tabular}[t]{p{6.2in}} 
\textbf{\textcolor{blue}{[Original Sentence]} Almost 1 million people were marooned by floodwater in about 10 districts in northern, northeastern and central parts of the country, as floodwater triggered by incessant monsoon rains have destroyed houses, submerged paddy fields and disrupted road transport in many places.}\\[1mm]
\toprule
$\bullet$ Almost 1 million people were marooned by \textcolor{orangered}{floodwater in about 10 districts in northern, northeastern and central parts of} the country, as floodwater triggered by incessant monsoon rains have destroyed houses, submerged paddy fields and disrupted road transport in many places. (2.8)\\[1mm]
$\bullet$ Almost 1 million people were marooned by floodwater in \textcolor{darkgreen}{about 10 districts in northern, northeastern and central parts of} the country, as floodwater triggered by incessant monsoon rains have destroyed houses, submerged paddy fields and disrupted road transport in many places. (3.2)\\[1mm]
$\bullet$ Almost 1 million people were marooned by floodwater in about 10 districts in northern, northeastern and central parts of the country, as \textcolor{dodgerblue}{floodwater triggered by incessant monsoon rains have} destroyed houses, submerged paddy fields and disrupted road transport in many places. (2.8)\\[1mm]
$\bullet$ Almost 1 million people were marooned by floodwater in about 10 districts in northern, northeastern and central parts of the country, as floodwater \textcolor{fuchsia}{triggered by incessant monsoon rains have destroyed houses}, submerged paddy fields and disrupted road transport in many places. (4.4)\\[1mm]
$\bullet$ Almost 1 million people were marooned by floodwater in about 10 districts in northern, northeastern and central parts of the country, as floodwater triggered by incessant monsoon rains have destroyed houses, \textcolor{saddlebrown}{submerged paddy fields and disrupted road transport in many places}. (4.0)\\[1mm]
\bottomrule
\end{tabular}}
\end{minipage}
\hfill
\end{scriptsize}
\caption{
Example text segments produced by the XLNet model. 
The scores of \emph{self-containedness} are shown in parentheses. 
Each segment is judged by five human evaluators on a scale of 1 (worst) to 5 (best) and we report their average scores. 
Human evaluation suggests that text segments generated by our model demonstrate a high degree of self-containedness.
} 
\label{tab:results-output}
\end{table*}

\begin{table*}[h!]
\setlength{\tabcolsep}{0pt}
\renewcommand{\arraystretch}{1.1}
\begin{scriptsize}
\begin{minipage}[b]{\hsize} 
\centering
\textsf{
\begin{tabular}[t]{p{6.2in}} 
\textbf{\textcolor{blue}{[Original Sentence]} It said the US States District Court for the Southern District of New York granted the application and appointed Irving H. Picard as trustee for the liquidation of the brokerage firm, while it named the law firm of Baker; Hostetler LLP as counsel to Picard.}\\[1mm]
\toprule
$\bullet$ It said the US States District Court for the Southern District of New York granted the application and appointed Irving H. Picard as \textcolor{orangered}{trustee for the liquidation of the brokerage firm}, while it named the law firm of Baker; Hostetler LLP as counsel to Picard. (3.6)\\[1mm]
$\bullet$ It said the US States District Court for the Southern District of New York \textcolor{darkgreen}{granted the application and appointed Irving H. Picard as trustee for} the liquidation of the brokerage firm, while it named the law firm of Baker; Hostetler LLP as counsel to Picard. (2.4)\\[1mm]
$\bullet$ It said the US States District Court for the Southern District of New York granted the application and appointed Irving H. Picard as trustee for the liquidation of the brokerage firm, \textcolor{dodgerblue}{while it named the law firm of Baker}; Hostetler LLP as counsel to Picard. (2.4)\\[1mm]
$\bullet$ It said the US States District Court for the Southern District of New York granted the application and appointed Irving H. Picard as trustee for the liquidation of the brokerage firm, while it named the law firm of Baker; \textcolor{fuchsia}{Hostetler LLP as counsel to Picard}. (2.4)\\[1mm]
\bottomrule
\end{tabular}}
\end{minipage}
\hfill
\end{scriptsize}
\caption{
Example text segments produced by the XLNet model. 
The scores of \emph{self-containedness} are shown in parentheses. 
Each segment is judged by five human evaluators on a scale of 1 (worst) to 5 (best) and we report their average scores. 
Human evaluation suggests that text segments generated by our model demonstrate a high degree of self-containedness.
} 
\label{tab:results-output}
\end{table*}

\begin{table*}[!h]
\setlength{\tabcolsep}{0pt}
\renewcommand{\arraystretch}{1.1}
\begin{scriptsize}
\begin{minipage}[b]{\hsize} 
\centering
\textsf{
\begin{tabular}[t]{p{6.2in}} 
\textbf{\textcolor{blue}{[Original Sentence]} But just in the last month, a so-called Floating Eyeballs toy made in China was recalled after it was found to be filled with kerosene, sets of toy drums and a toy bear were also recalled because of lead paint and an infant wrist rattle was recalled because of a choking hazard.}\\[1mm]
\toprule
$\bullet$ \textcolor{orangered}{But just in the last month}, a so-called Floating Eyeballs toy made in China was recalled after it was found to be filled with kerosene, sets of toy drums and a toy bear were also recalled because of lead paint and an infant wrist rattle was recalled because of a choking hazard. (2.4)\\[1mm]
$\bullet$ But just in the last month, a so-called Floating Eyeballs toy made in China was recalled \textcolor{darkgreen}{after it was found to} be filled with kerosene, sets of toy drums and a toy bear were also recalled because of lead paint and an infant wrist rattle was recalled because of a choking hazard. (1.8)\\[1mm]
$\bullet$ But just in the last month, a so-called Floating Eyeballs toy made in \textcolor{dodgerblue}{China was recalled after it was found to be filled with} kerosene, sets of toy drums and a toy bear were also recalled because of lead paint and an infant wrist rattle was recalled because of a choking hazard. (2.6)\\[1mm]
$\bullet$ But just in the last month, a so-called Floating Eyeballs toy made in China was recalled after it was found to be filled with kerosene, sets \textcolor{fuchsia}{of toy drums and a toy bear were} also recalled because of lead paint and an infant wrist rattle was recalled because of a choking hazard. (1.4)\\[1mm]
$\bullet$ But just in the last month, a so-called Floating Eyeballs toy made in China was recalled after it was found to be filled with kerosene, sets \textcolor{saddlebrown}{of toy drums and a toy bear were also} recalled because of lead paint and an infant wrist rattle was recalled because of a choking hazard. (2.0)\\[1mm]
\bottomrule
\end{tabular}}
\end{minipage}
\hfill
\end{scriptsize}
\caption{
Example text segments produced by the XLNet model. 
The scores of \emph{self-containedness} are shown in parentheses. 
Each segment is judged by five human evaluators on a scale of 1 (worst) to 5 (best) and we report their average scores. 
This example is among the worst cases; we use it to illustrate the difficulty of finding self-contained segments in a polynomial space. 
}
\label{tab:results-output}
\vspace{-0.1in}
\end{table*}

\end{document}